\documentclass{article} 
\usepackage{iclr2015,times}
\usepackage{hyperref}
\usepackage{url}
\usepackage{graphicx}
\usepackage{bm}
\usepackage{amsmath}
\usepackage{amsfonts}
\usepackage{array}

\def\vtheta{{\bm{\theta}}}

\def\vu{{\bm{u}}}
\def\vv{{\bm{v}}}

\def\mH{{\bm{H}}}

\title{Qualitatively Characterizing Neural\\Network Optimization Problems}

\author{
  Ian J. Goodfellow$^*$ \& Oriol Vinyals$^*$ \& Andrew M. Saxe$^{**}$ \\
$^*$Google Inc., Mountain View, CA\\
$^{**}$Department of Electrical Engineering, Stanford University, Stanford, CA\\
\texttt{\{goodfellow,vinyals\}@google.com}, \texttt{asaxe@stanford.edu}
}

\iclrfinalcopy 

\iclrconference 

\begin{document}

\maketitle

\begin{abstract}
Training neural networks involves solving large-scale non-convex optimization problems.
This task has long been believed to be extremely difficult, with fear of local minima
and other obstacles motivating a variety of schemes to improve optimization, such
as unsupervised pretraining. However, modern neural networks are able to achieve negligible
training error on complex tasks, using only direct training with stochastic gradient
descent. We introduce a simple analysis technique to look for evidence that such networks
are overcoming local optima. We find that, in fact, on a straight path from initialization
to solution, a variety of state of the art neural networks never encounter any significant
obstacles.
\end{abstract}

\section{Introduction}
Neural networks are generally regarded as difficult to optimize.
The objective functions we must optimize in order to train them
are non-convex and there are not many theoretical guarantees
about the performance of the most popular algorithms on these
problems. Nevertheless, neural networks are commonly trained
successfully and obtain state of the art results on many tasks.

In this paper, we present a variety of simple experiments
designed to roughly characterize the objective functions
involved in neural network training. These experiments are
not intended to measure any specific quantitative property
of the objective function, but rather to answer some simple
qualitative questions. Do neural networks enter and escape
a series of local minima? Do they move at varying speed
as they approach and then pass a variety of saddle points?

Answering these questions definitively is difficult, but we
present evidence strongly suggesting that the answer to
all of these questions is no. We show that there exists
a linear subspace in which neural network training could
proceed by descending a single smooth slope with no barriers.
Early symmetry breaking is the most
conspicuous example of non-convexity. 

One important question is what happens after SGD leaves this
well-behaved linear subspace.
The main text of this article is restricted to experiments
that were peer-reviewed prior to ICLR 2014, but the
appendix presents additional experiments added after the review
process ended. These experiments show that in some cases SGD
does encounter obstacles, such as a ravine that shapes its path,
but we never found evidence that local minima or saddle points
slowed the SGD trajectory.
This suggests that less exotic problems
such as poor conditioning and variance in the gradient estimate
are the primary difficulties in training neural networks.

In all cases, we examine the total cost function (added up
across all training examples). SGD of course only ever
acts on unbiased stochastic approximations to this loss
function. The structure of these stochastic approximations
could be different from the global loss
functions that we examine here, so it remains possible that
neural networks are difficult to train due to exotic structures
in individual terms of the total cost function, or due to the
noise induced by sampling minibatches of these terms.

The results of our linear subspace experiments were qualitatively the same for
all seven models we examined, which were drawn from a variety of categories, including
fully-connected supervised feed-forward networks
~\citep{Rumelhart86c}
with a variety of activation functions,
supervised convolutional networks~\citep{lecun2010convolutional},
unsupervised models, recurrent models of sequences,
and analytically tractable factored linear models.
(The additional experiments in the appendix found some qualitatively
unique behavior outside this linear subspace for two of our models, but the remainder
have the same qualitative behavior as factored linear models)

Our models were all chosen because they performed well on
competitive benchmark tasks. More research is needed to
determine whether one should interpret our results as
implying that SGD never encounters exotic obstacles when training
neural networks, or as implying that SGD only works well when
it does not encounter these structures.

\section{Linear path experiments}

Training a neural network consists of finding the optimal set of parameters
$\vtheta$. These are initialized to some set of small, random, initial parameters $\vtheta = \vtheta_i$. We then train using
stochastic gradient descent (usually with extra features such as momentum) 
to minimize $J(\vtheta)$ until
reaching convergence (usually some early stopping criterion).
At the end of training, $\vtheta = \vtheta_f$.

The trajectory that SGD follows from $\vtheta_i$ to $\vtheta_f$ is complicated and
high-dimensional. It is difficult to summarize such a trajectory meaningfully in
a two-dimensional visualization. Simple learning curves showing the value of the
objective function over time do not convey some fairly simple information. For
example, when a learning curve bounces up and down repeatedly, we do not know
whether the objective function is highly bumpy or whether SGD is rapidly changing
direction due to noise in the stochastic, minibatch-based, estimate of the gradient.
When the objective function remains constant for long periods of time, we do not
know whether the parameters are stuck in a flat region, oscillating around a local
minimum, or tracing their way around the perimeter of a large obstacle.

In this paper, we introduce a simple technique for qualitatively analyzing objective
functions. We simply evaluate $J(\vtheta)$ at a series of points
$\vtheta = (1 - \alpha) \vtheta_0 + \alpha \vtheta_1$
for varying values of $\alpha$. This sweeps out a line in parameter space. We can
see whether the cross-section of the objective function along this line is well-behaved.

When we set $\vtheta_0 = \vtheta_i$ and
$\vtheta_1=\vtheta_f$, we find that the objective function
has a simple, approximately convex shape along this cross-section. In other words,
if we knew the correct direction, a single coarse line search
could do a good job of training a neural network.

These results are consistent with recent empirical and theoretical work arguing that local minima are not a significant
problem for training large neural networks
~\citep{Saxe-et-al-ICLR13,saddle,saddle_nyu}.

\section{Feed-forward fully connected networks}

\begin{figure}
\centering
\begin{tabular}{cc}
    \includegraphics[width=.4\textwidth]{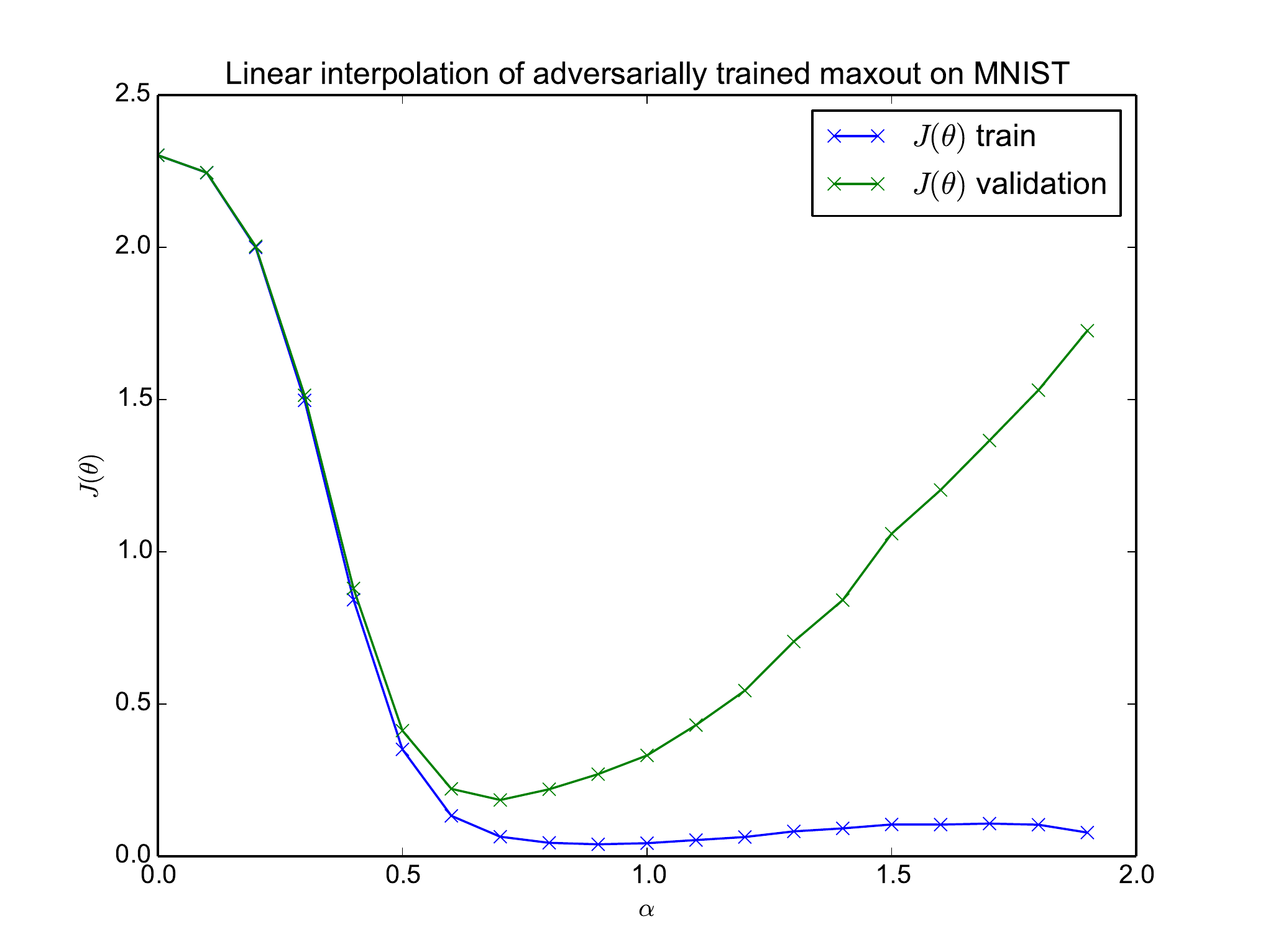} &
    \includegraphics[width=.4\textwidth]{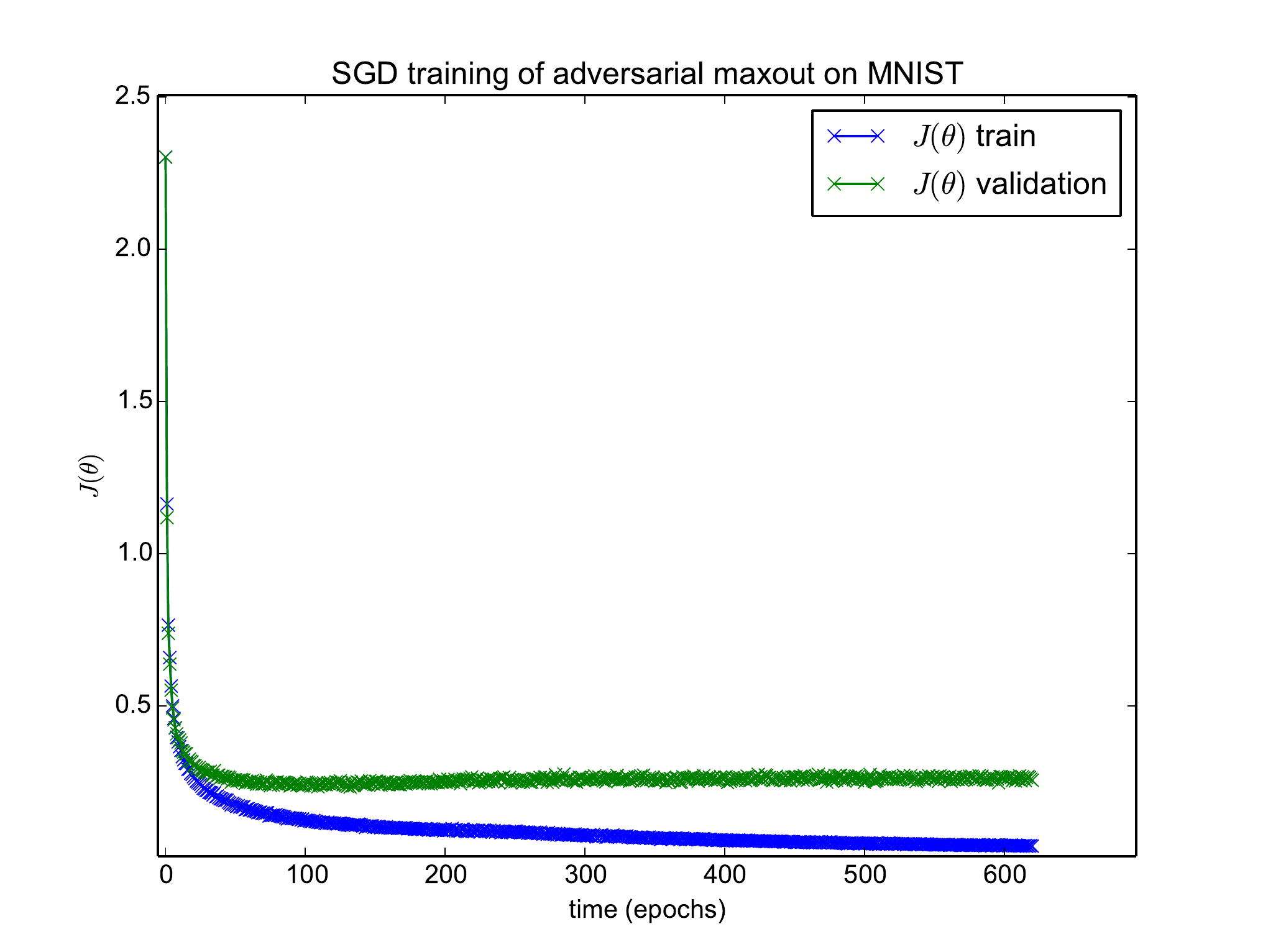} \\
    \includegraphics[width=.4\textwidth]{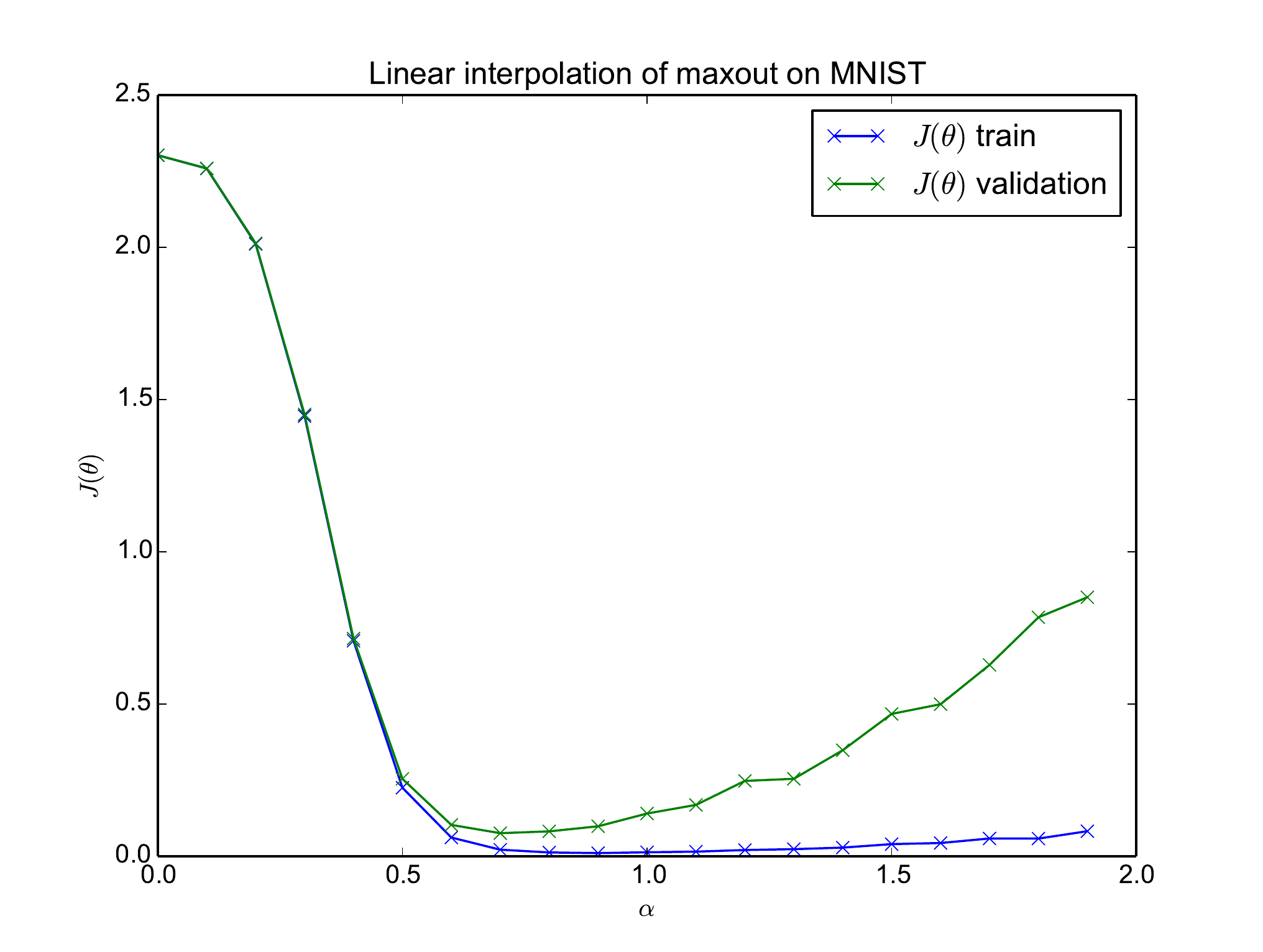} &
    \includegraphics[width=.4\textwidth]{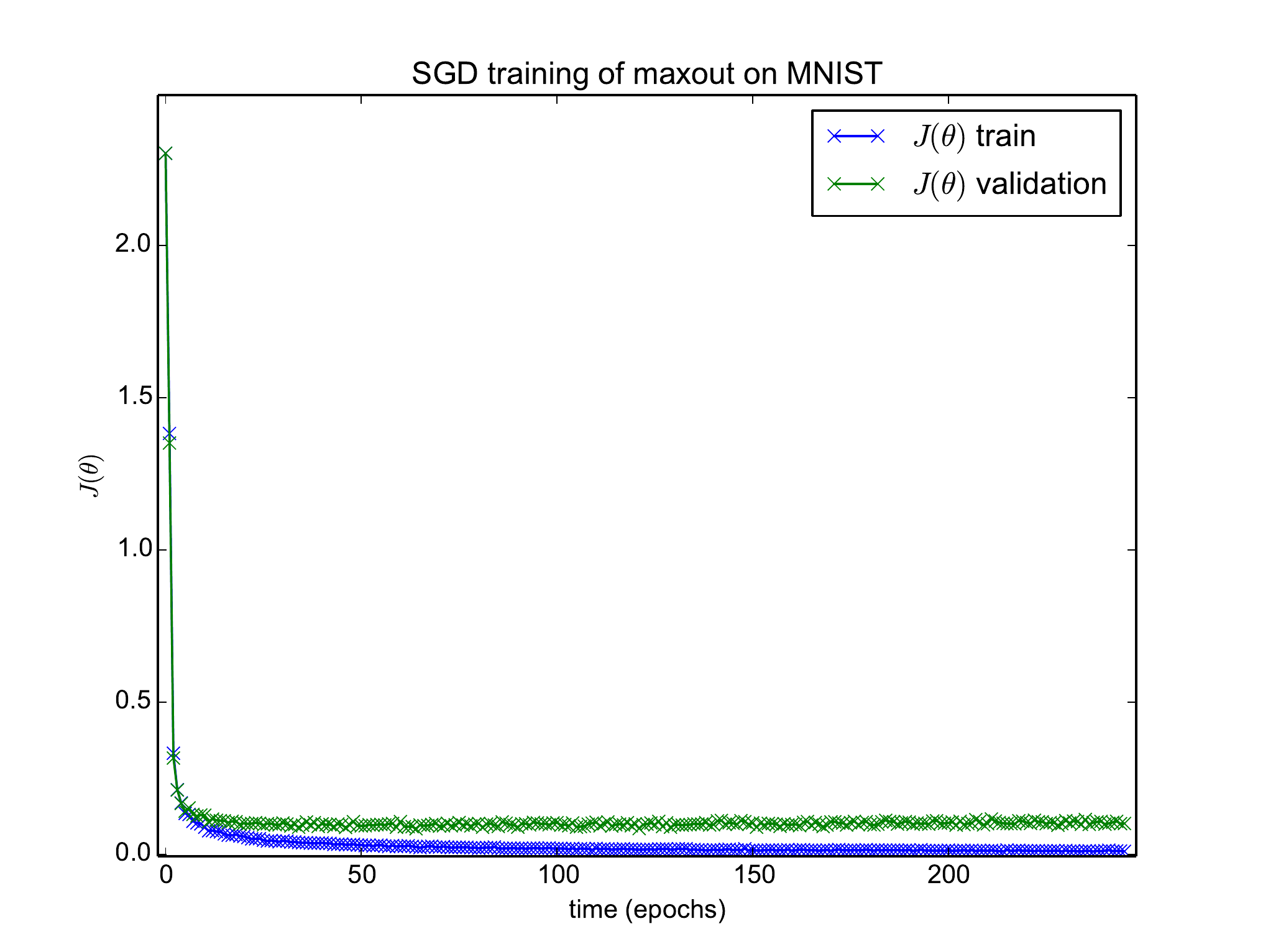}
\end{tabular}
\caption{
Experiments with maxout on MNIST.
Top row) The state of the art model, with adversarial training.
Bottom row) The previous best maxout network, without adversarial training.
Left column) The linear interpolation experiment. This experiment shows that
the objective function is fairly smooth within the 1-D subspace spanning
the initial and final parameters of the model. Apart from the flattening
near $\alpha = 0$, it appears nearly convex in this subspace. If we chose
the initial direction correctly, we could solve the problem with a coarse
line search.
Right column) The progress of the actual SGD algorithm over time. The vast majority
of learning happens in the first few epochs. Thereafter, the algorithm
struggles to make progress.
}
\label{maxout}
\end{figure}

\begin{figure}
\centering
\begin{tabular}{cc}
    \includegraphics[width=.4 \textwidth]{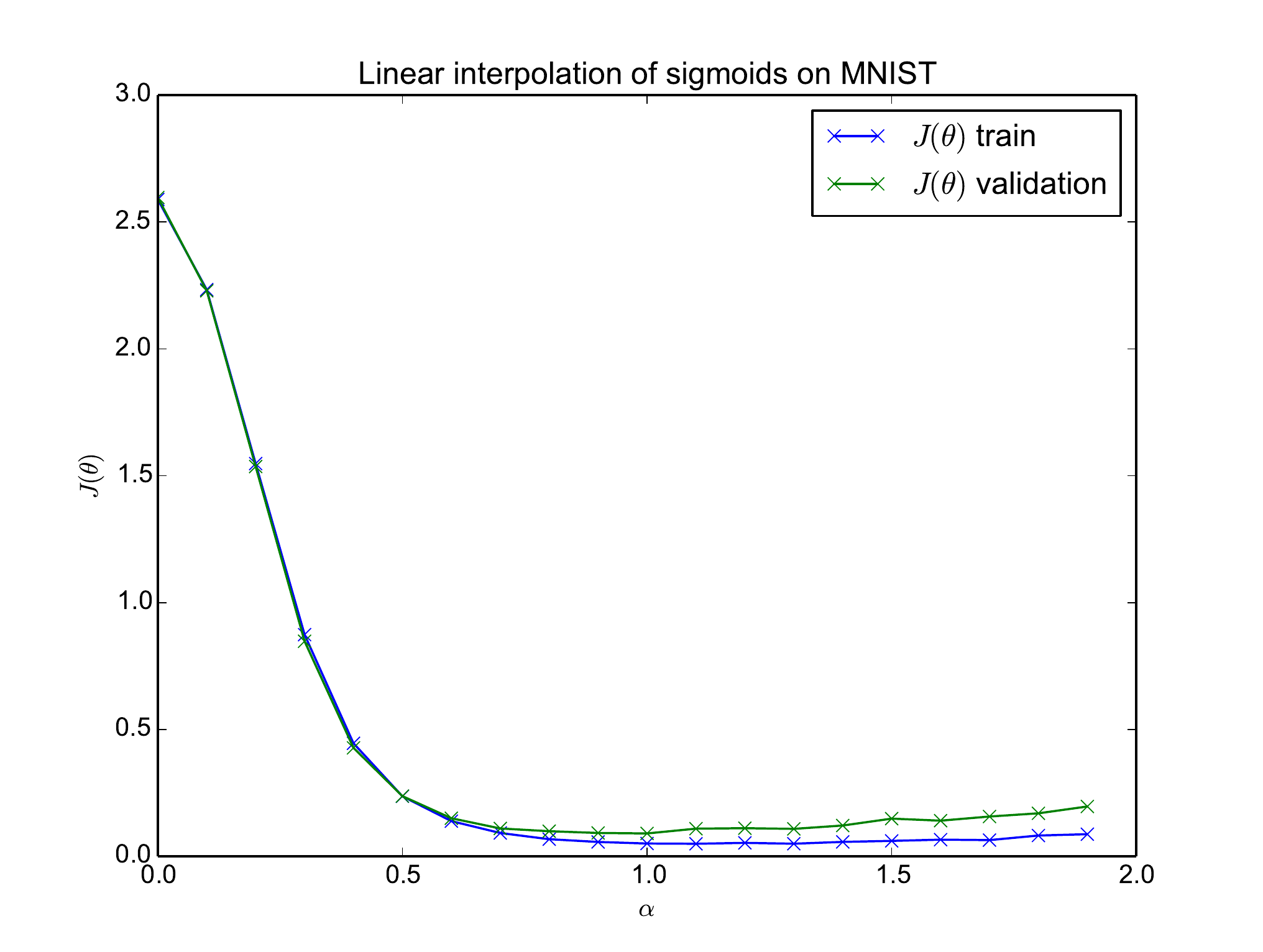} &
\includegraphics[width=.4 \textwidth]{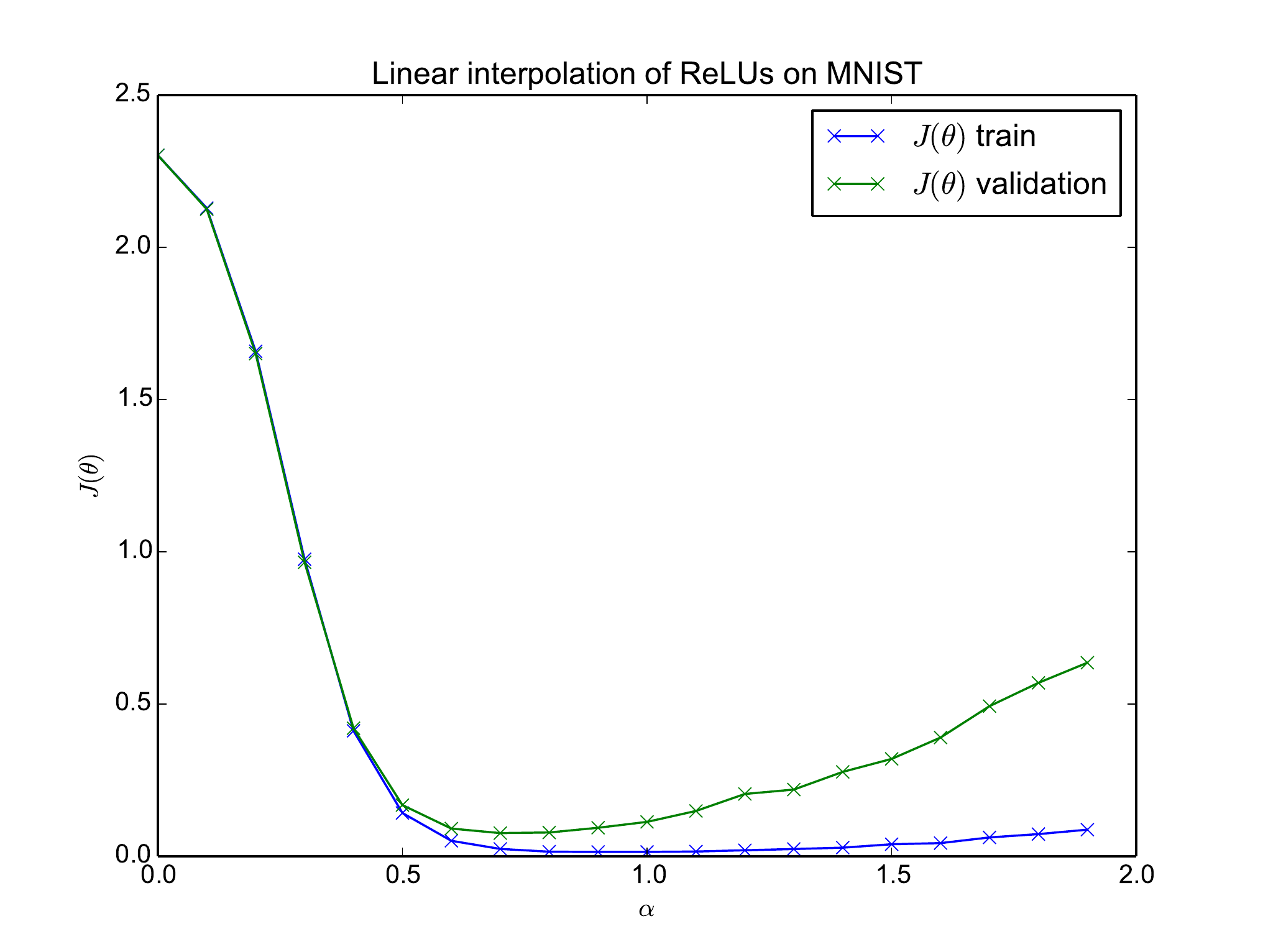}
\end{tabular}
\caption{The linear interpolation curves for fully connected networks
with different activation functions. Left) Sigmoid activation function.
Right) ReLU activation function.}
\label{fig:other_act}
\end{figure}

\begin{figure}
\centering
\includegraphics[width=.4 \textwidth]{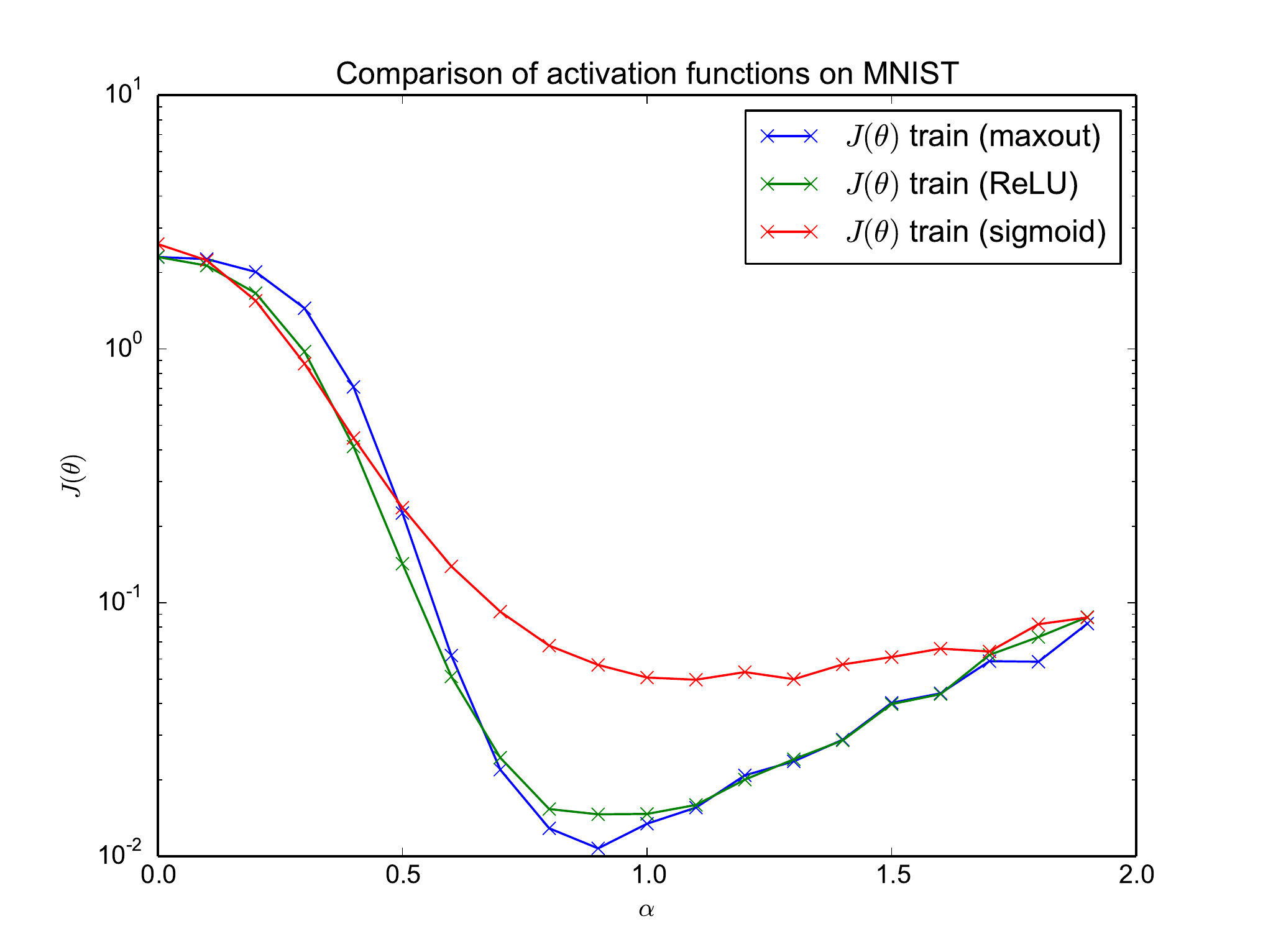}
\caption{The linear interpolation experiment for maxout, ReLUs, and sigmoids
on MNIST, all plotted on the same axis for comparison. For this plot, we put
the y axis in log scale, otherwise differences at the bottom of the curve are
difficult to see.}
\label{fig:comparison}
\end{figure}

We begin our investigation with the simplest kind of neural network, the deterministic,
feed-forward, fully-connected supervised network. For these experiments we use the MNIST dataset~\citep{LeCun+98}.
When not using dataset augmentation, the best result in this category is a maxout network~\citep{Goodfellow-et-al-ICML2013}
regularized with dropout~\citep{dropout} and adversarial training~\citep{Goodfellow-AdvTrain2015},
and trained using SGD with momentum.
See the appendix of this paper for a full specification of the architecture and training algorithm
for this and all subsequent experiments.
This configuration results in an average of 78.2 mistakes on the MNIST test set, out of 10,000 examples total.
Without adversarial training, the model also performs very well, with only 94 mistakes.
Running the linear interpolation experiment on this
problem, we find in Fig.~\ref{maxout} that the 1-D subspace spanning the initial parameters
and final parameters is very well-behaved, and that SGD spends most of its time exploring
the flat region at the bottom of the valley.
Maxout units do not saturate (they can saturate with
respect to their input, but not with respect to their parameters), so perhaps it should not be too surprising that optimization is simple in
this case. To determine whether the hard zero saturation of ReLUs~\citep{Jarrett-ICCV2009,Glorot+al-AI-2011}
or the two-sided saturation of logistic sigmoids can induce additional difficulties, we ran
the linear interpolation experiment for both of these activation functions. The results
are presented in Fig.~\ref{fig:other_act} and Fig.~\ref{fig:comparison}. Again, we find that the 1-D subspace spanning
the initial and final parameters contains no difficult, exotic structures.

\begin{figure}
\centering
\begin{tabular}{cc}
    \includegraphics[width=.4\textwidth]{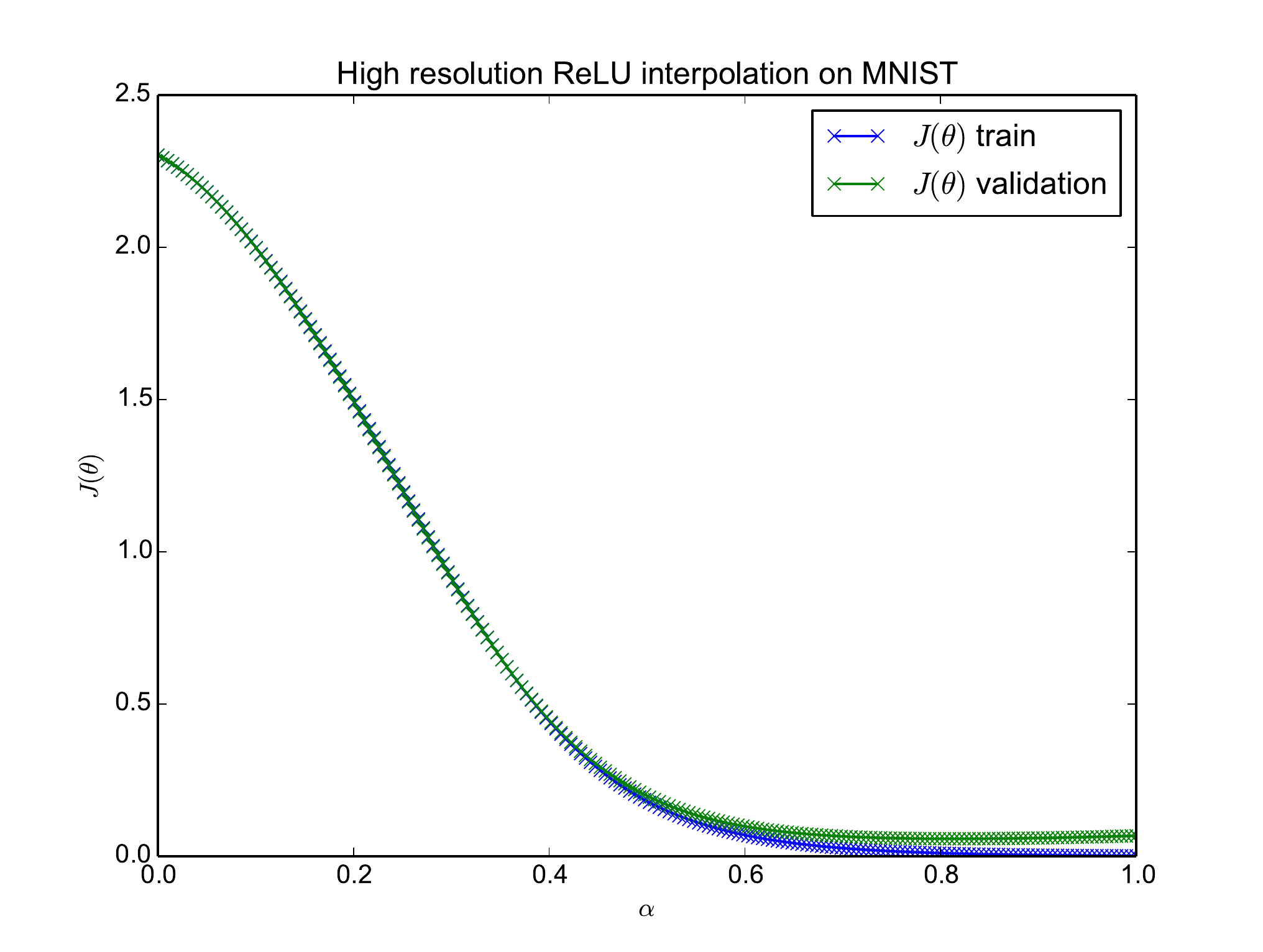} &
    \includegraphics[width=.4\textwidth]{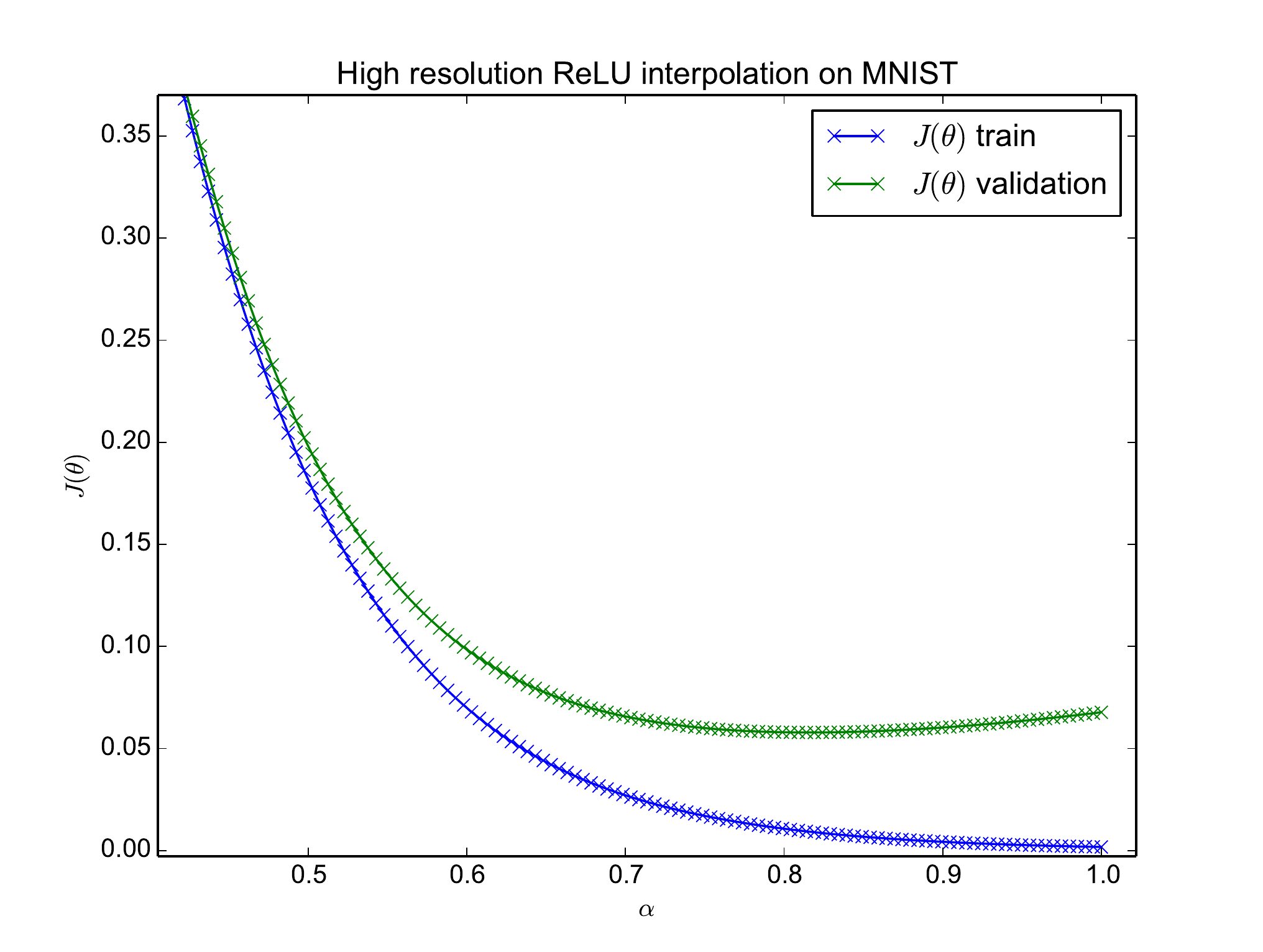} \\
    (a) & (b) \\
    \includegraphics[width=.4\textwidth]{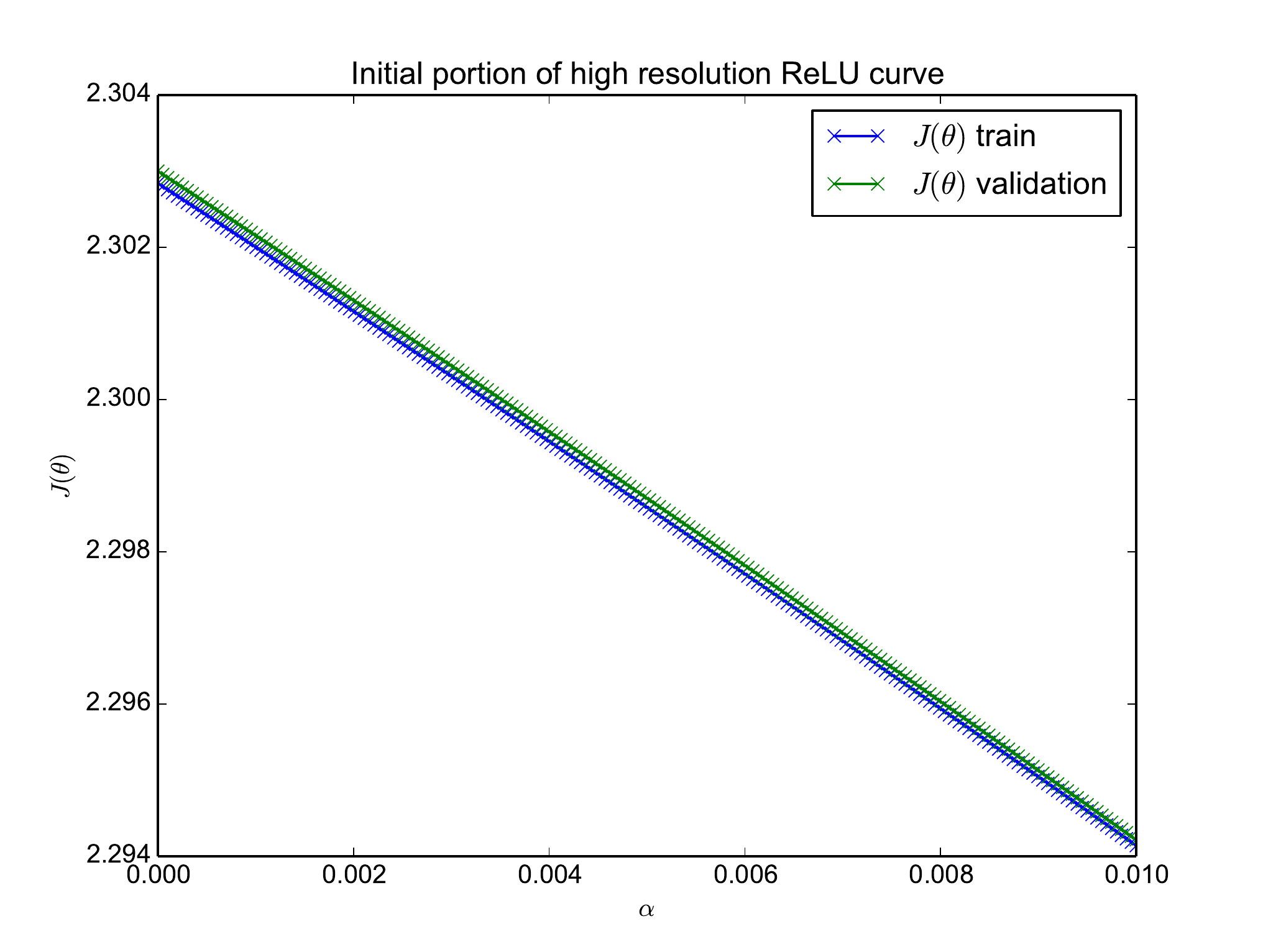} &
    \includegraphics[width=.4\textwidth]{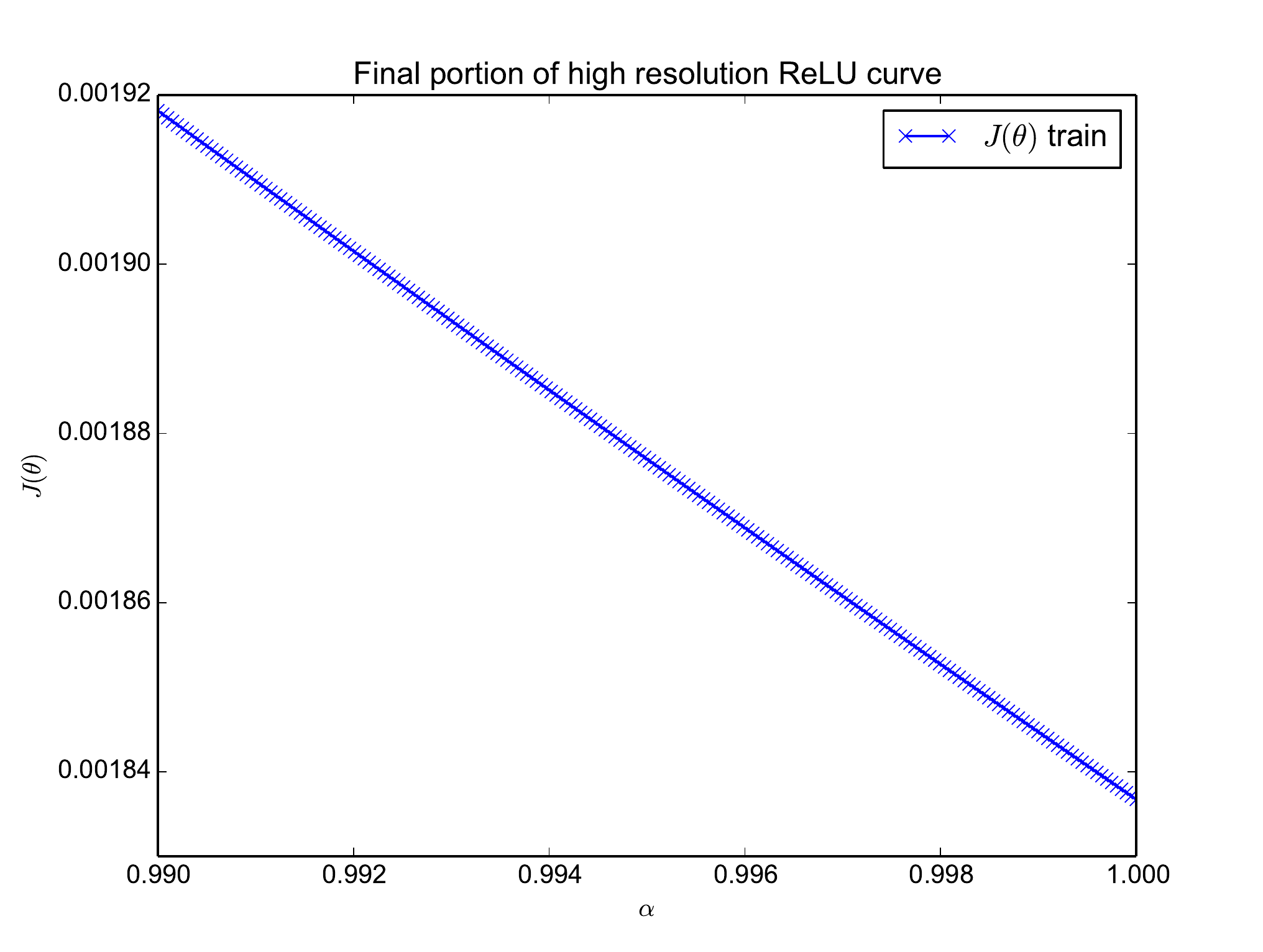}\\
    (c) & (d)
\end{tabular}
\caption{Higher resolution linear interpolation experiments.
    a) Tiling the interval [0, 1] with 200 values of $\alpha$.
    b) A zoomed-in view of the same plot.
    c) Tiling the interval [0, .01] with 200 values of $\alpha$, to see
    whether the initial symmetry breaking causes difficult structures.
    d) Tiling the interval [.99, 1.] with 200 values of $\alpha$, to see
    if the behavior of the objective function is more exotic in regions
    where the parameters encode fully learned intermediate concepts.
    We do not show the validation set objective here because it is too
    widely separated from the training set objective and would require
    zooming out the plot too far.
}
\label{fig:fine}
\end{figure}

One possible objection to these results is that we have explored $\alpha$ with too
coarse of a resolution to expose local non-convex structures. We therefore ran
a variety of higher-resolution experiments, presented in Fig.~\ref{fig:fine}.
For these experiments, we did not use dropout, because the resolution we use
here is high enough to expose artifacts induced by the Monte Carlo approximation
to the true dropout loss function, which involves a sum over all (exponentially many)
dropout masks. Maxout tends to overfit on MNIST if used without dropout, so
we used ReLUs for these experiments.
We found that increased resolution
did not expose any small, difficult structures.

\begin{figure}
\centering
\begin{tabular}{cc}
    \includegraphics[width=.4\textwidth]{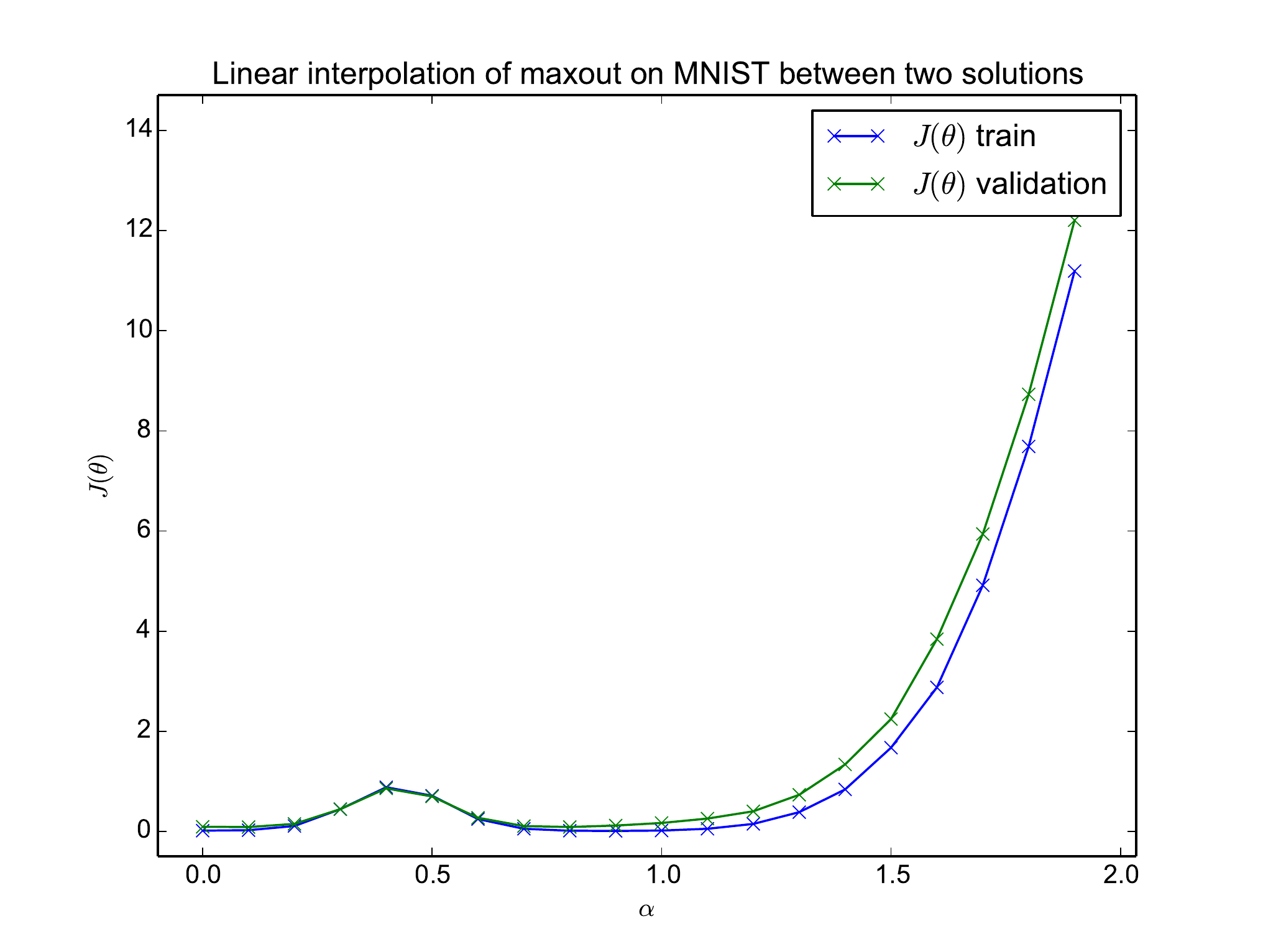} &
\includegraphics[width=.4\textwidth]{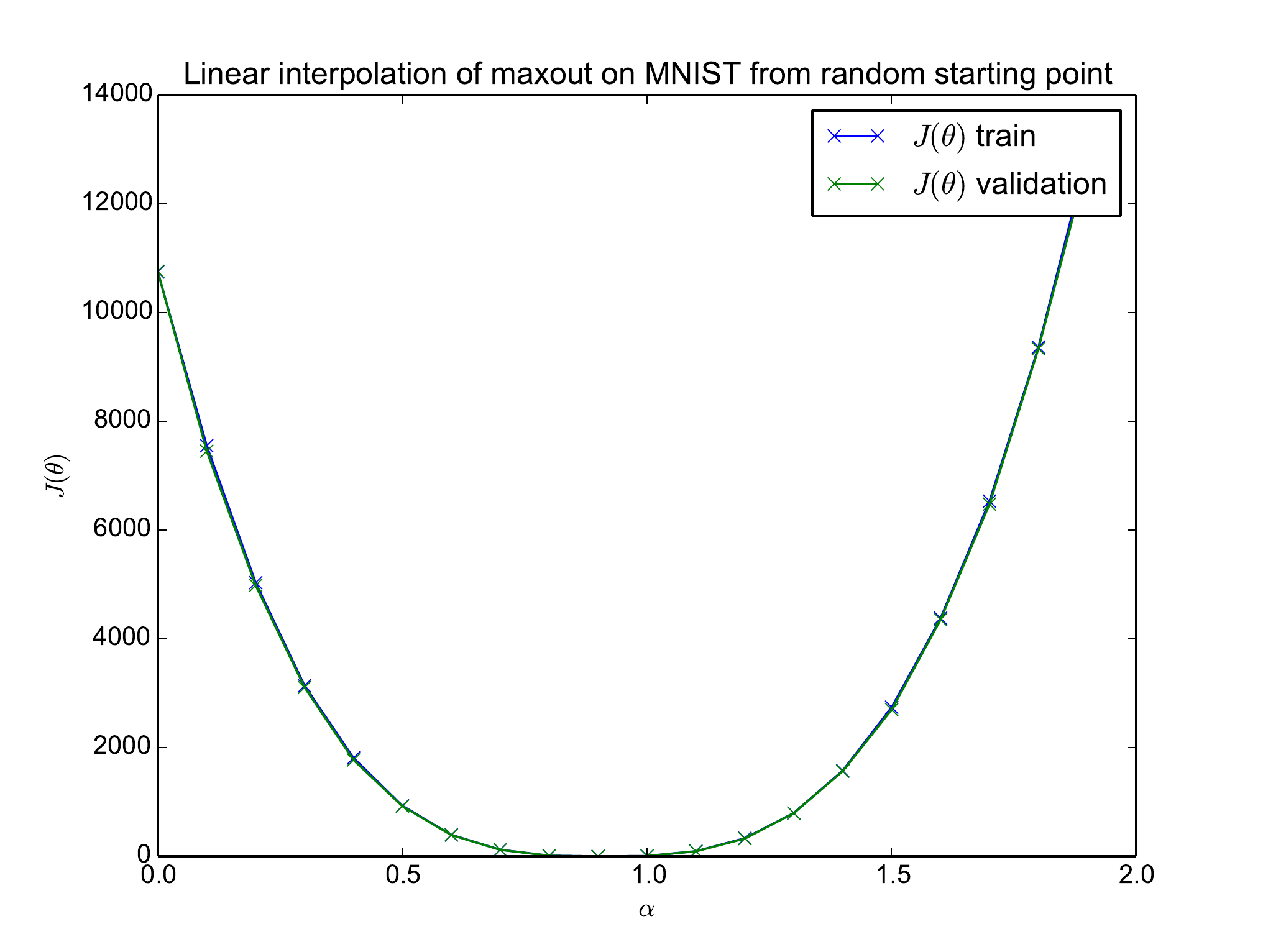}
\end{tabular}
\caption{Here we use linear interpolation to search for local minima. Left)
By interpolating between two different SGD solutions, we show that each solution
is a different local minimum within this 1-D subspace.
Right) If we interpolate between a random point in space and an SGD solution,
we find no local minima besides the SGD solution, suggesting that local minima are rare.
}
\label{fig:local}
\end{figure}

There are of course multiple minima in neural network optimization problems,
and the shortest path between two minima can contain a barrier of higher cost.
We can find two different solutions by using different random seeds for the
random number generators used to initialize the weights, generate dropout masks,
and select examples for SGD minibatches. (It is possible that these solutions
are not minima but saddle points that SGD failed to escape)
We do not find any local minima within this subspace other than solution points,
and these different solutions appear to correspond to different choices of how
to break the symmetry of the saddle point at the origin, rather than to fundamentally
different solutions of varying quality.
See Fig.~\ref{fig:local}.

\section{Advanced networks}

Having performed experiments to understand the behavior of neural network optimization
on supervised feedforward networks, we now verify that the same behavior occurs
for more advanced networks.

\begin{figure}
    \centering
    \begin{tabular}{cc}
        \includegraphics[width=.4\textwidth]{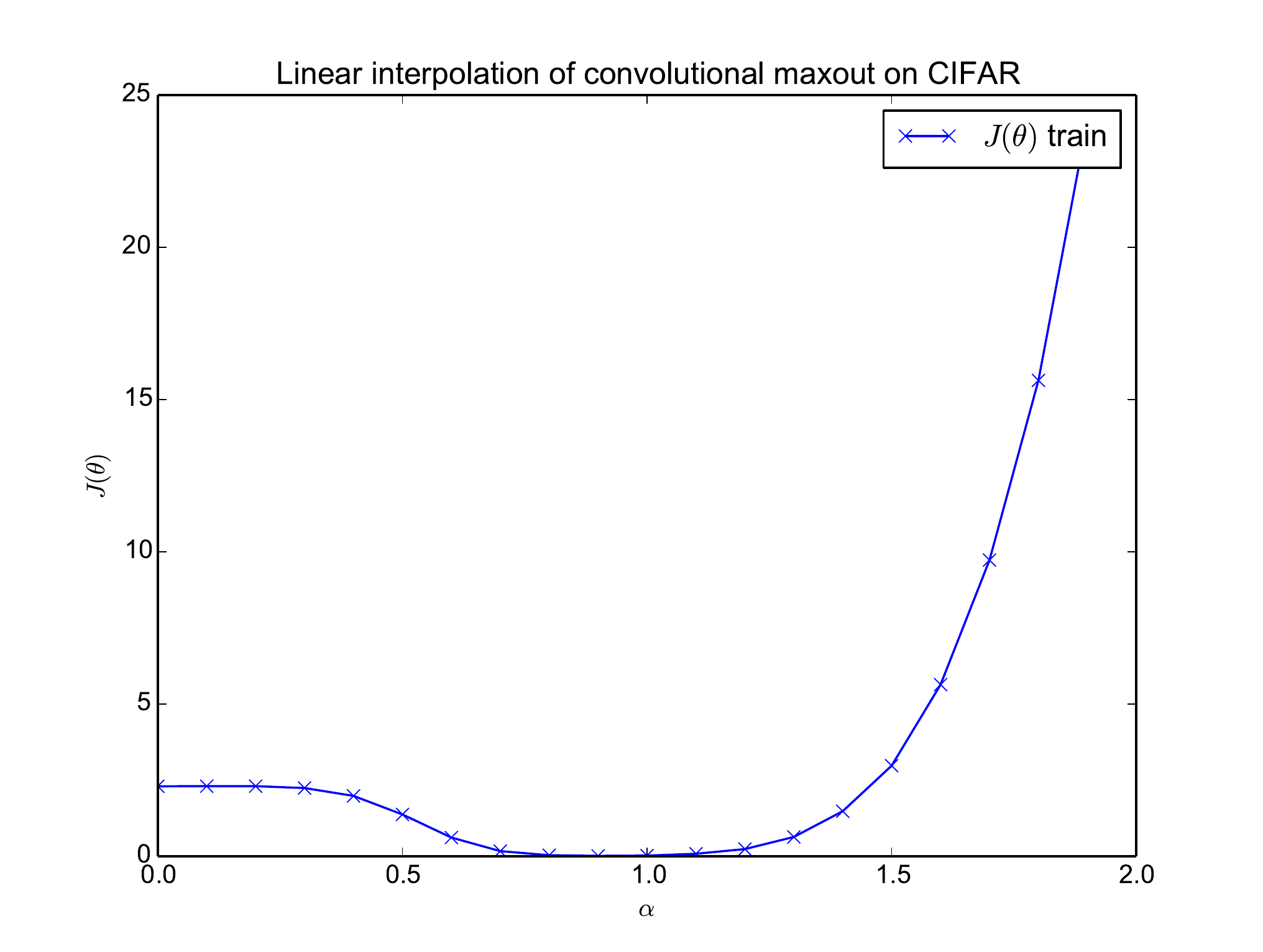} &
        \includegraphics[width=.4\textwidth]{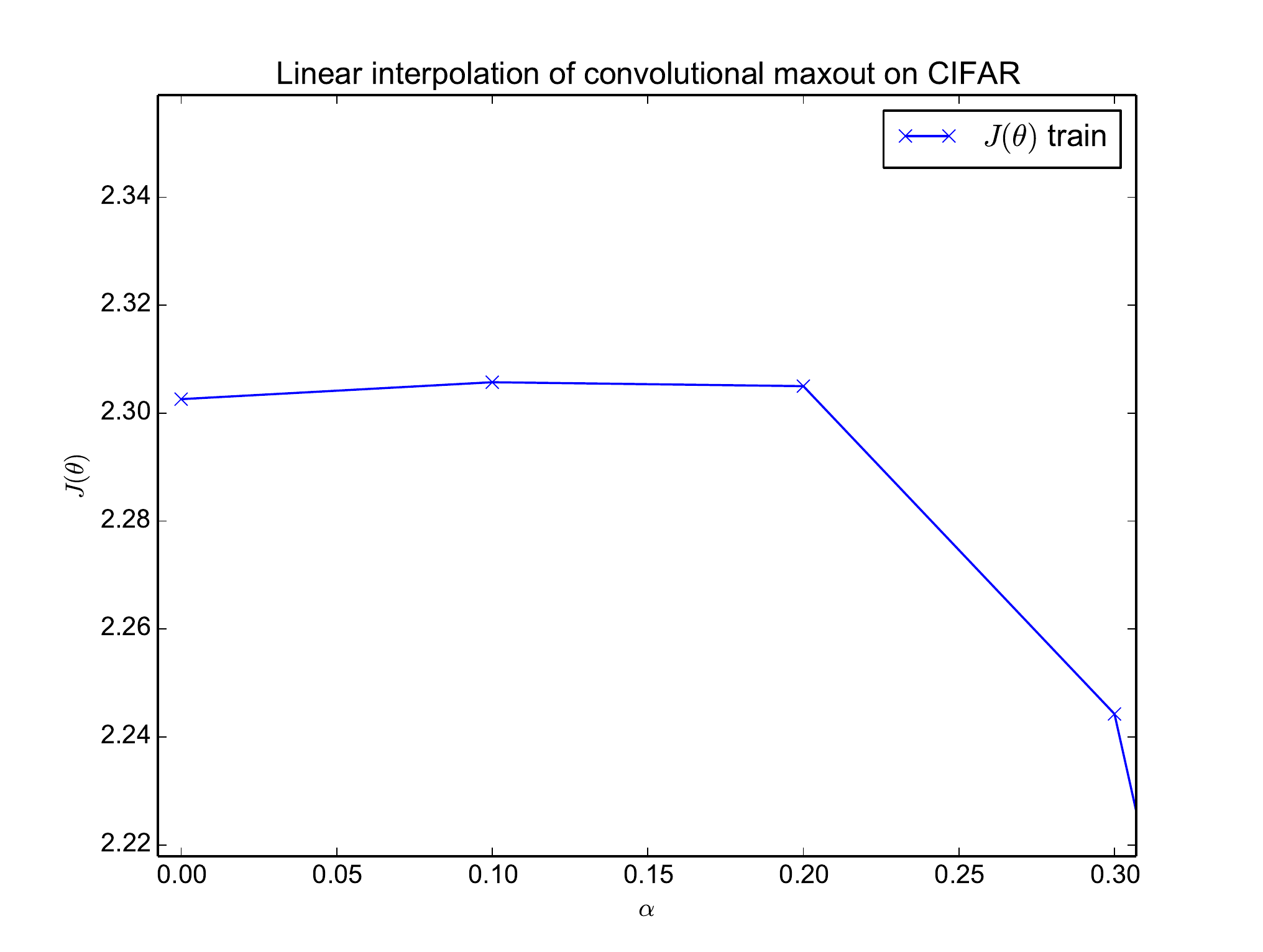}
    \end{tabular}
    \caption{The linear interpolation experiment for a convolutional maxout network
        on the CIFAR-10 dataset~\citep{KrizhevskyHinton2009}. Left) At a global scale,
        the curve looks very well-behaved. Right) Zoomed in near the initial point,
        we see there is a shallow barrier that SGD must navigate.
    }
    \label{convnet}
\end{figure}

In the case of convolutional networks, we find that there is a single barrier in the
objective function, near where the network is initialized. This may simply correspond
to the network being initialized with too large of random weights. This barrier is
reasonably wide but not very tall. See Fig.~\ref{convnet} for details.

\begin{figure}
\centering
\begin{tabular}{cc}
\includegraphics[width=.4 \textwidth]{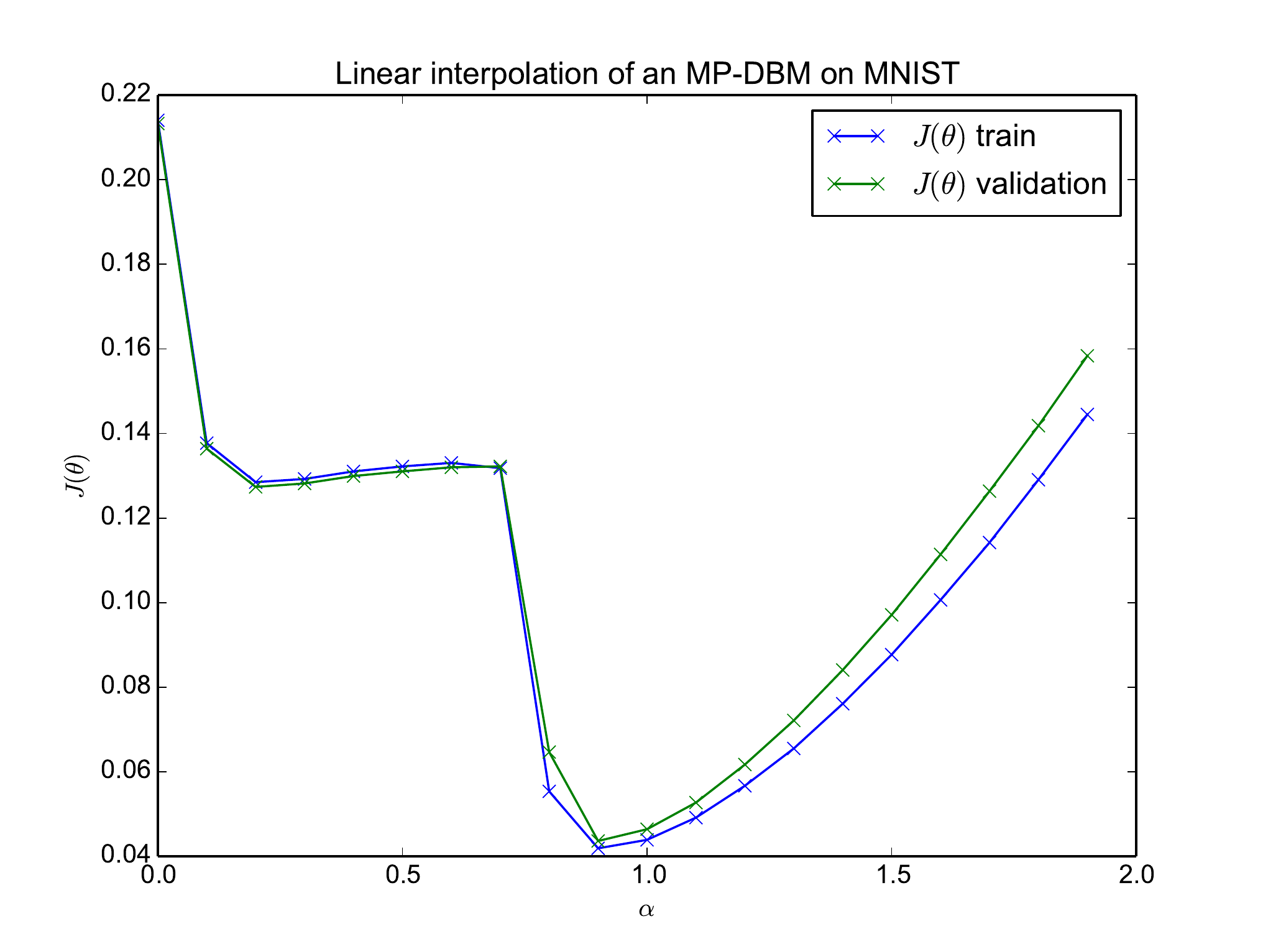} &
\includegraphics[width=.4 \textwidth]{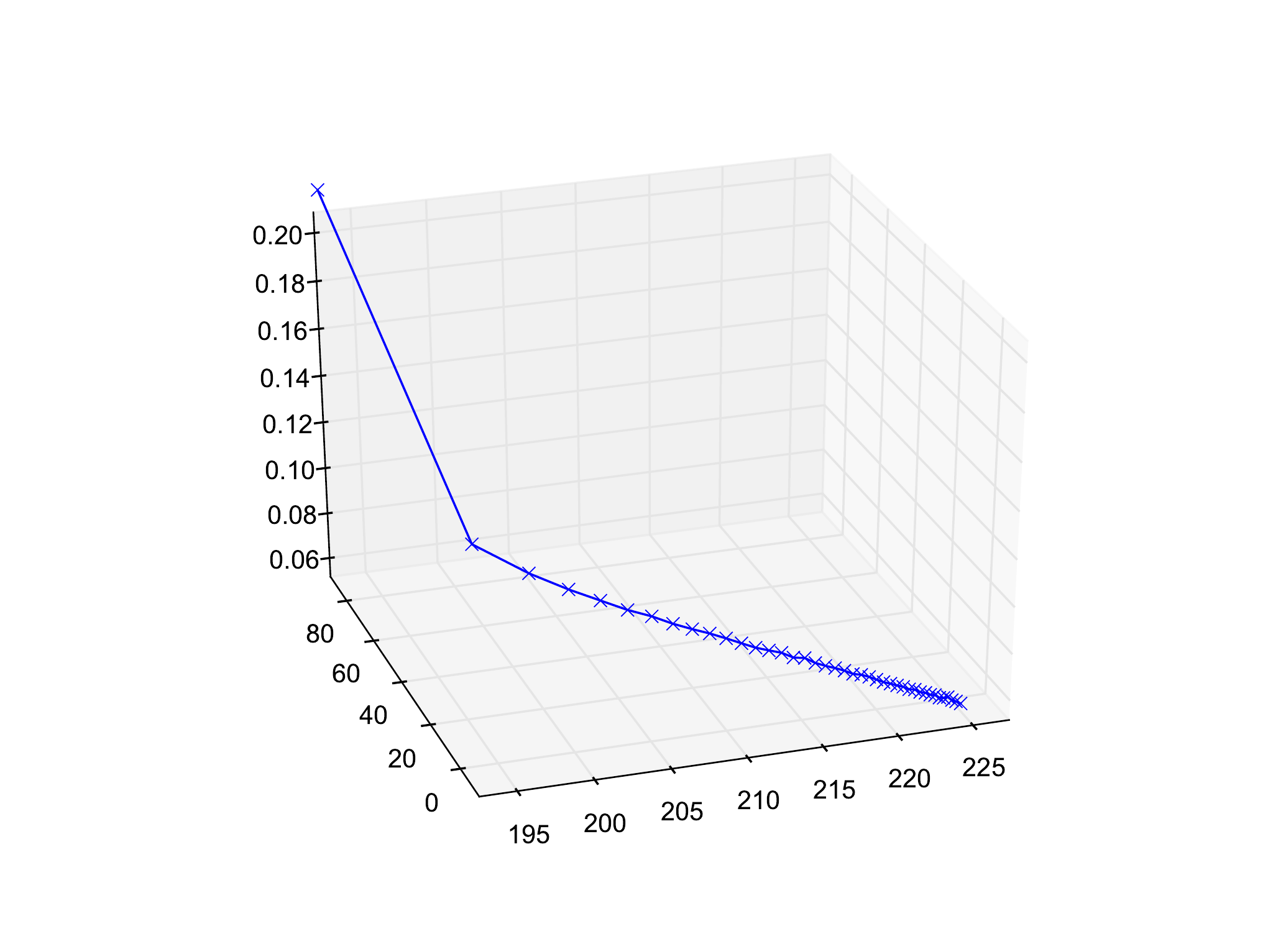}
\end{tabular}
\caption{Experiments with the MP-DBM. Left) The linear interpolation experiment reveals a secondary local minimum with high error.
Right) On the two horizonal axes, we plot components of $\theta$ that capture the extrema of $\theta$
throughout the learning process. On the vertical axis, we plot the objective function. Each point is
another epoch of actual SGD learning. This plot allows us to see that SGD did not pass near this anomaly.}
\label{fig:mpdbm}
\end{figure}

To examine the behavior of SGD on generative models, we experimented with an MP-DBM~\citep{mpdbm}.
This model is useful for our purposes because it gets good performance as a generative model
and as a classifier, and its objective function is easy to evaluate (no MCMC business).
Here we find a secondary local minimum with high error, but a visualization of the SGD trajectory
reveals that SGD passed far enough around the anomaly to avoid having it affect learning.
See Fig.~\ref{fig:mpdbm}.
The MP-DBM was initialized with very large, sparse, weights, which may have contributed to our
visualization technique exploring more non-convex areas, e.g. due to saturation of sigmoidal units.

\begin{figure}
\centering
\includegraphics[width=.4 \textwidth]{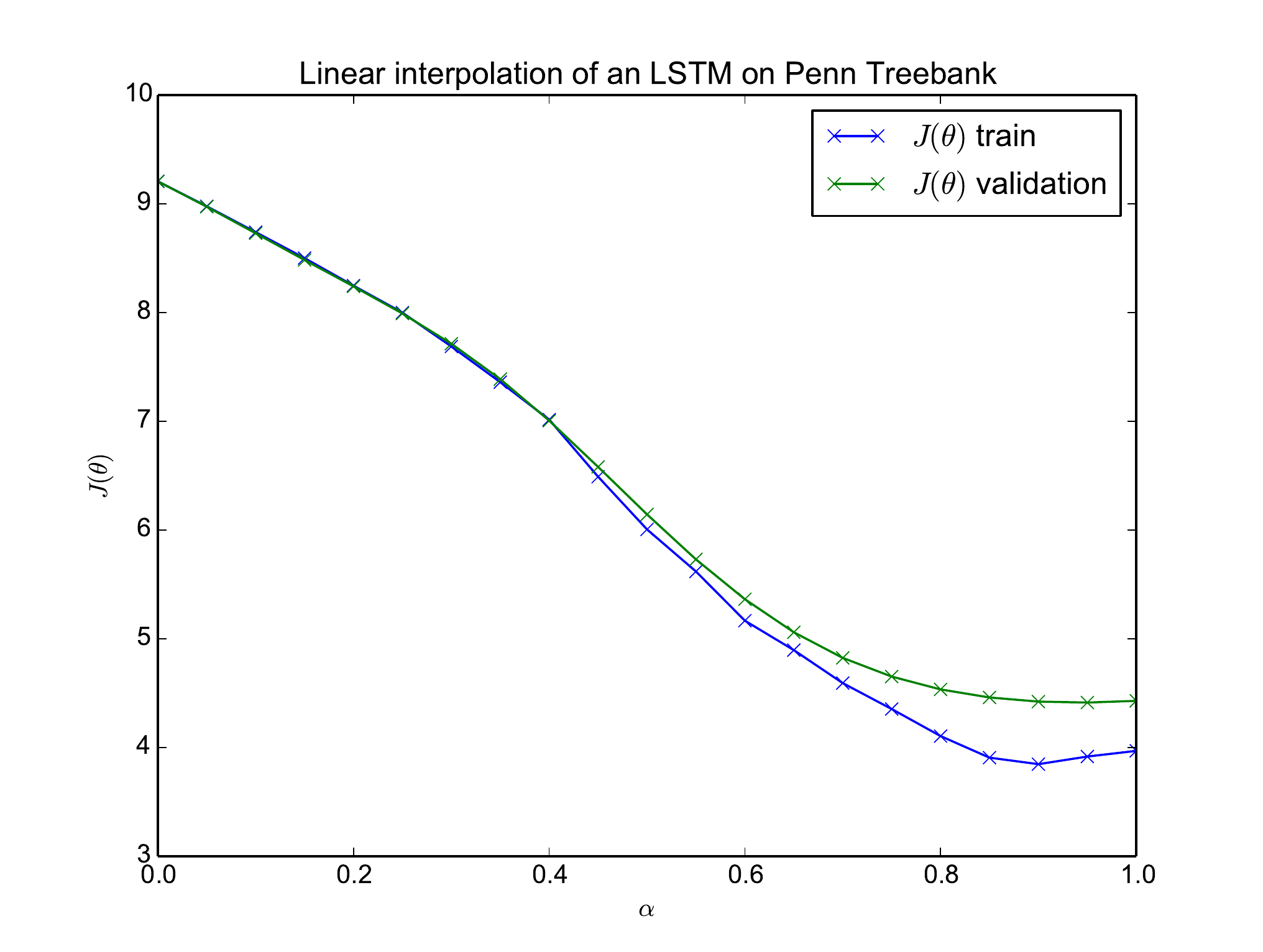}
\caption{The linear interpolation experiment for an LSTM trained on the Penn Treebank dataset.}
\label{fig:treebank}
\end{figure}

Finally, we performed the linear interpolation experiment for an LSTM
regularized with dropout~\citep{lstm,dropout_lstm}
on the Penn Treebank dataset~\citep{Marcus93buildinga}. See Fig.~\ref{fig:treebank}.
This experiment did not find any difficult structures, showing that the exotic
features of non-convex optimization do not appear to cause difficulty even for 
recurrent models of sequences.

\section{Deep linear networks}

~\citet{Saxe-et-al-ICLR13} have advocated developing a mathematical theory of
deep networks by studying simplified mathematical models of these networks.
Deep networks are formed by composing an alternating series of learned affine transformations
and fixed non-linearities. One simplified way to model these functions is to compose
only a series of learned linear transformations. The composition of a series
of linear transformations is itself a linear transformation, so this mathematical model
lacks the expressive capacity of a general deep network. However, because the
weights of such a model are factored, its learning dynamics resemble those of the
deep network. The output of the model is linear in the input to the model, but non-linear
as a function of the model parameters.
In particular, while fitting linear regression parameters is a convex problem, 
fitting deep linear regression parameters is a non-convex problem.

Deep linear regression suffers from saddle points but does not suffer from local
minima of varying quality. All minima are global minima, and are linked to
each other in a continuous manifold.

Our linear interpolation experiments can be carried out {\em analytically}
rather than experimentally in the case of deep linear regression. The results are
strikingly similar to our results with deep non-linear networks.

\begin{figure}
  \centering
  \includegraphics[width=.4\textwidth]{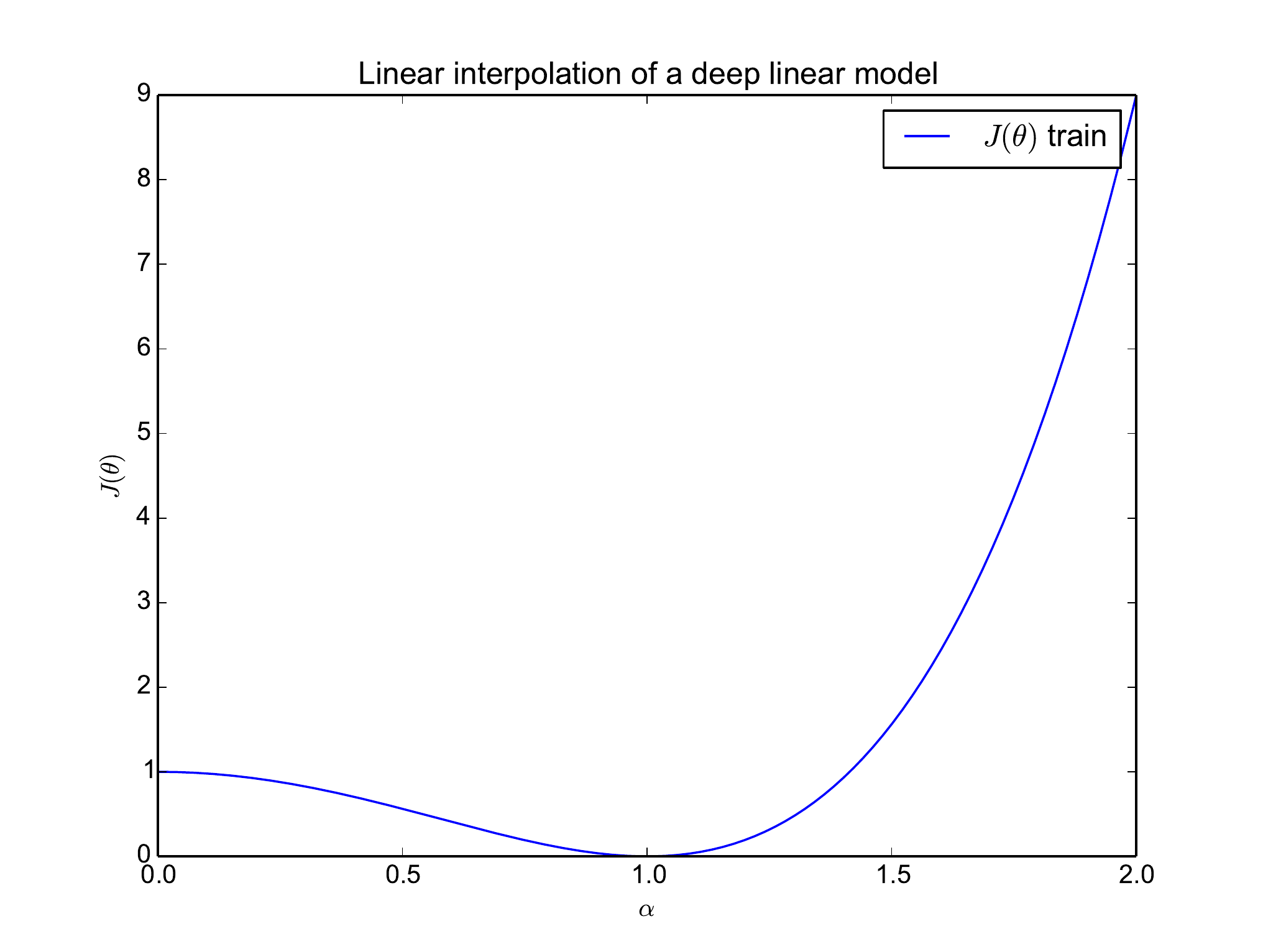}
\caption{
Linear interpolation from a small random initialization point
to a solution for a linear regression model of depth 2. This shows
the same qualitative features as our linear interpolation experiments
for neural networks: a flattening of the objective function near the
saddle point at the origin, and only one minimum within this 1-D
subspace.
}
\label{fig:linear_plot}
\end{figure}

\begin{figure}
\centering
\begin{tabular}{cc}
\includegraphics[width=.4\textwidth]{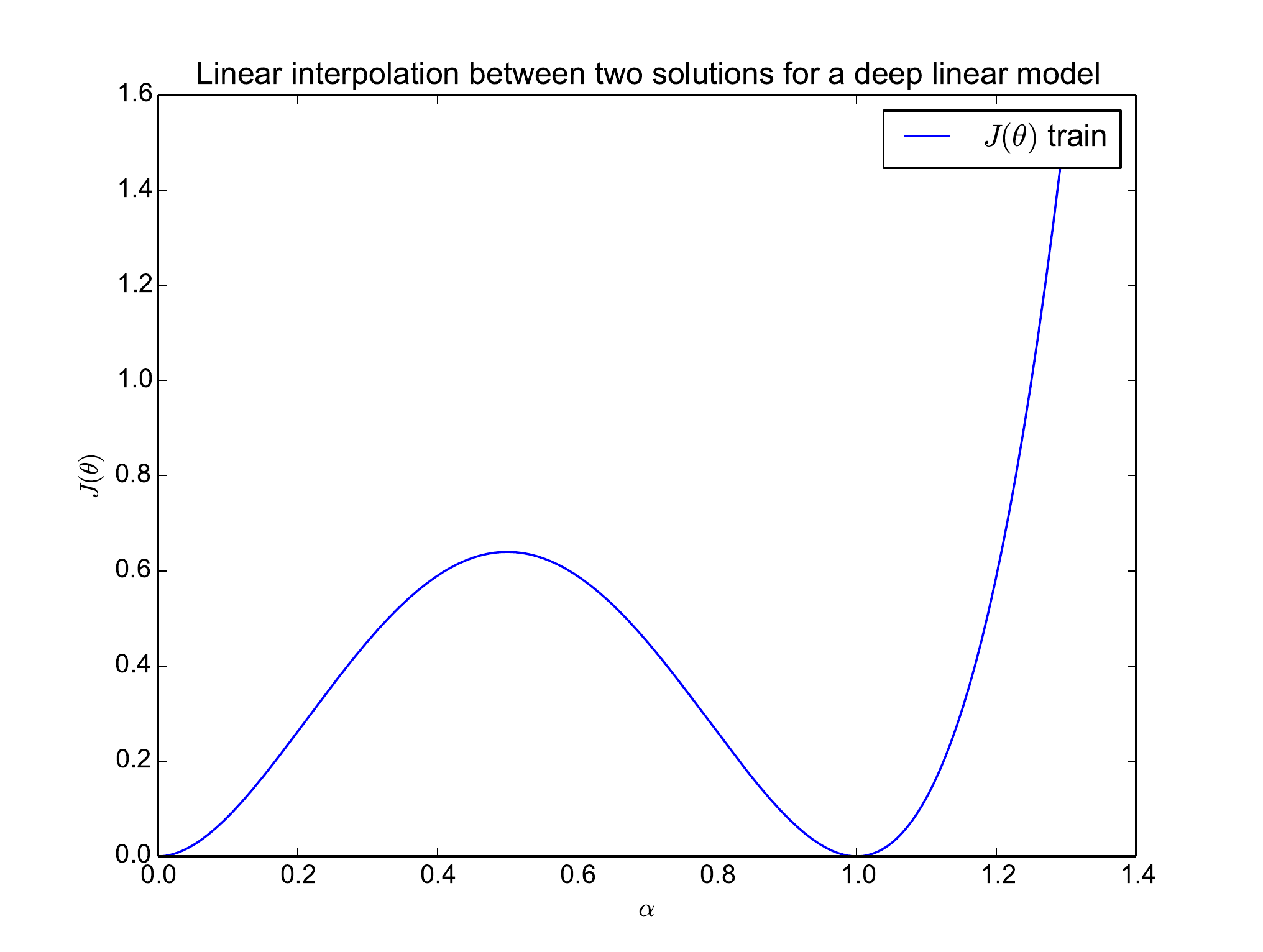} & %
  \includegraphics[width=.4\textwidth]{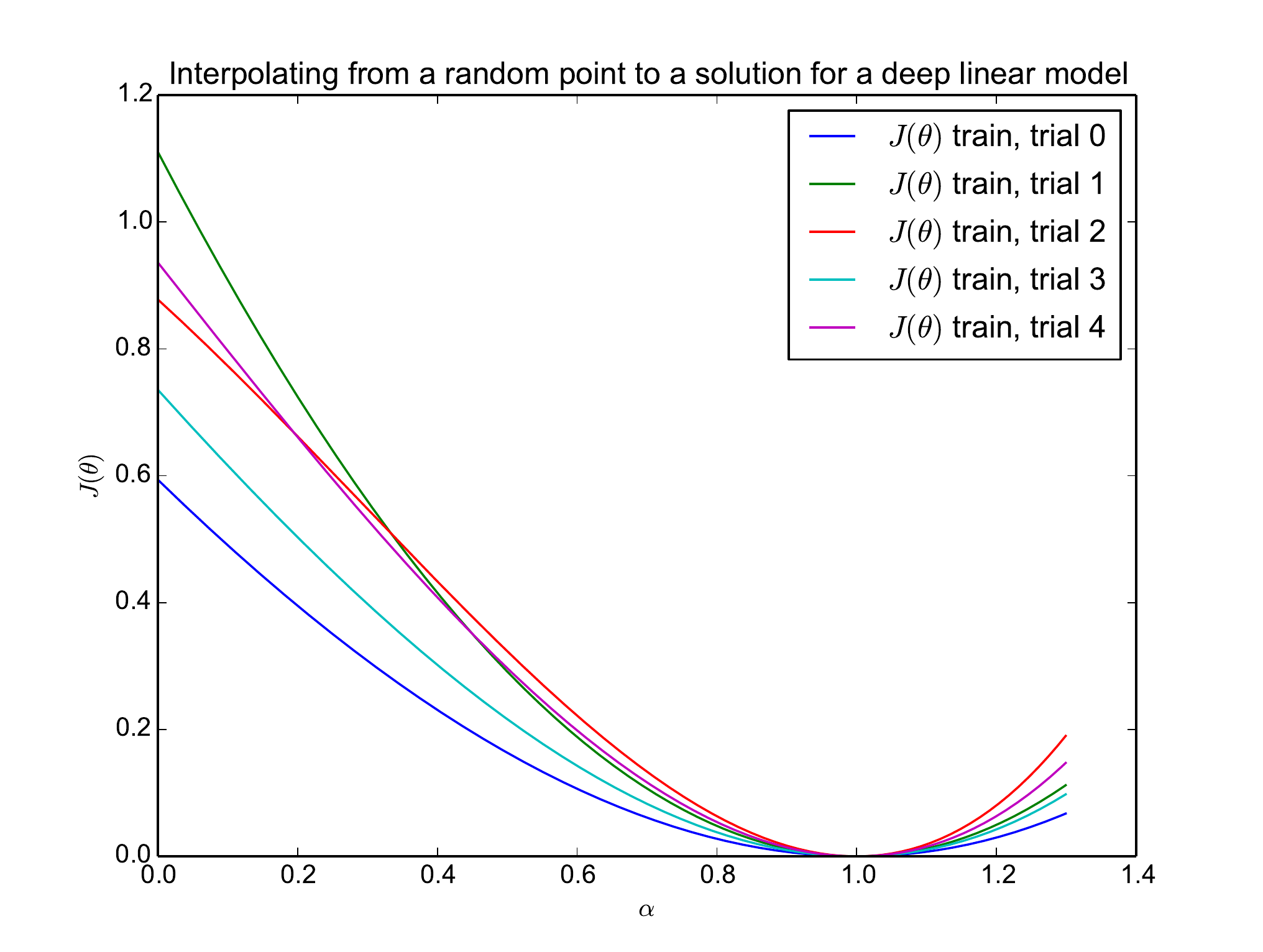}
\end{tabular}
\caption{Left) Interpolation between two solutions to deep linear
regression. Though these two solutions lie on connected manifold
of globally minimal values, the straight line between them encounters
a barrier of higher cost. The curve for the low dimensional linear model
has all the same qualitative characteristics as the curve for the high
dimensional non-linear networks we studied.
Right) Interpolation between a random point with large norm and
an solution to deep linear regression. As with the neural network,
this search does not encounter any minima other than the solution
used to initialize the search.
}
\label{fig:linear_two_point}
\end{figure}

Specifically, we show that the problem of training $y = w_1 w_2 x$ to output $1$
when $x=1$ using mean squared error is sufficient to produce all of the qualitative
features of neural network training that our linear interpolation experiments have
exposed. See Fig.~\ref{fig:linear_plot} and Fig.~\ref{fig:linear_two_point}.

\section{Discussion}

The reason for the success of SGD on a wide variety of tasks is now clear: these
tasks are relatively easy to optimize. Finding a good direction in a high-dimensional
space is a difficult problem, but it is not nearly as difficult as navigating an error
surface that has complicated obstacles within multiple different low-dimensional subspaces.

This work has only considered neural networks that perform very well. It is possible
that these neural networks perform well because extensive hyperparameter search has
found problems that SGD is able to optimize easily, but that other hyperparameters
correspond to optimization problems that are too hard. In particular, it seems likely
that very large neural networks are easier to fit to a particular task.

Future work should aim to characterize the set of problems that are easy for SGD,
to understand why SGD is able to avoid the obstacles that are present, and to
determine why the training of large models remains slow despite the scarcity of obstacles.
More advanced optimization algorithms could reduce the computational cost of deploying
neural networks by enabling small networks to reach good performance, and could reduce
the cost of training large networks by reducing the amount of time required to reach
their solution.

\subsubsection*{Acknowledgments}

We would like to thank J\"org Bornschein, Eric Drexler, and Yann Dauphin for helpful discussions.
We would like to thank Yaroslav Bulatov, Chung-Cheng Chiu, Greg
Corrado, and Jeff Dean for their reviews of drafts of this work.
We would like to thank the developers of Theano\citep{bergstra+al:2010-scipy,Bastien-Theano-2012}
and Pylearn2\citep{pylearn2_arxiv_2013}.

\bibliography{iclr2015}
\bibliographystyle{iclr2015}

\appendix

\section{Experiment details}

All of our experiments except for the sigmoid network were using hyperparameters taken
directly from the literature. We fully specify each of them here.

{\em Adversarially trained maxout network}: This model is the one used by ~\citet{Goodfellow-AdvTrain2015}.
There is no public configuration for it, but the paper describes how to modify the previous best maxout network to obtain it.

{\em Maxout network}: This model was retrained using the publicly available implementation
used by~\citet{Goodfellow-et-al-ICML2013}. The code is available at:

\url{https://github.com/lisa-lab/pylearn2/blob/master/pylearn2/scripts/papers/maxout/mnist_pi.yaml}

{\em ReLU network with dropout}: This model is intended to nearly reproduce the
the dropout ReLU network described by ~\citet{DropoutThesis}.
It is a standard reference implementation provided by Pylearn2~\citep{pylearn2_arxiv_2013} and the specific file is available here:

\url{https://github.com/lisa-lab/pylearn2/blob/master/pylearn2/scripts/papers/dropout/mnist_valid.yaml}

{\em ReLU network without dropout}: We simply removed the dropout from the preceding configuration file.

{\em Sigmoid network}: We simply replaced the ReLU non-linearities with sigmoids. This performs acceptably
for a sigmoid network; it gets a test set error rate of 1.66\%.

{\em Convolutional network}: We used the best convolutional network available in Pylearn2
for the CIFAR-10 dataset, specifically the one developed by ~\citet{Goodfellow-et-al-ICML2013}.
In order to reduce the computational cost of computing the training set objective, we used
the the variant without data augmentation. The configuration file is available here:
\url{https://github.com/lisa-lab/pylearn2/blob/master/pylearn2/scripts/papers/maxout/cifar10_no_aug_valid.yaml}

{\em MP-DBM}: We used the best MP-DBM for classifying MNIST, as described by~\citet{mpdbm}.

{\em Dropout LSTM}: We used the configuration described in the paper introducing this method~\citep{dropout_lstm}.

\section{Straying from the path}

We have shown that, for seven different models of practical interest, there exists
a straight path from initialization to solution along which the objective function
decreases smoothly and monotonically, at least up to the resolution that our
experiments investigated.

Stochastic gradient descent does not actually follow this path. We know that SGD
matches this path at the beginning and at the end.

One might naturally want to plot the norm of the residual of the parameter value
after projecting the parameters at each point in time into the 1-D subspace we have
identified. SGD begins at $\vtheta_i$, the solution point chosen
by early stopping is $\vtheta_f$, and SGD visits $\vtheta(t)$ at time $t$.
If we define $\vu$ to be a unit vector pointing in the direction $\vtheta_f - \vtheta_i$,
then the primary subspace we have investigated so far is the line
$\vtheta_i + \alpha(t) \vu$ where $\alpha(t) = (\vtheta(t) - \vtheta_i)^\top \vu$.
We can plot the coordinate within this subspace, $\alpha(t)$,
on the horizontal axis. We can then define a second unit vector $\vv$ pointed in the
direction $\vtheta(t) - (\vtheta_i + \alpha(t) \vu)$. The projection into this subspace
is $\beta(t) = (\vtheta(t) - \vtheta_i - \alpha(t) \vu)^\top \vv$. In other words,
$\beta(t)$ is the norm of the residual of the projection of $\vtheta_i$ onto the line
spanning initialization and solution. We can plot $\beta$ on the vertical axis.

\begin{figure}
  \centering
  \begin{tabular}{cc}
    \includegraphics[width=.4\textwidth]{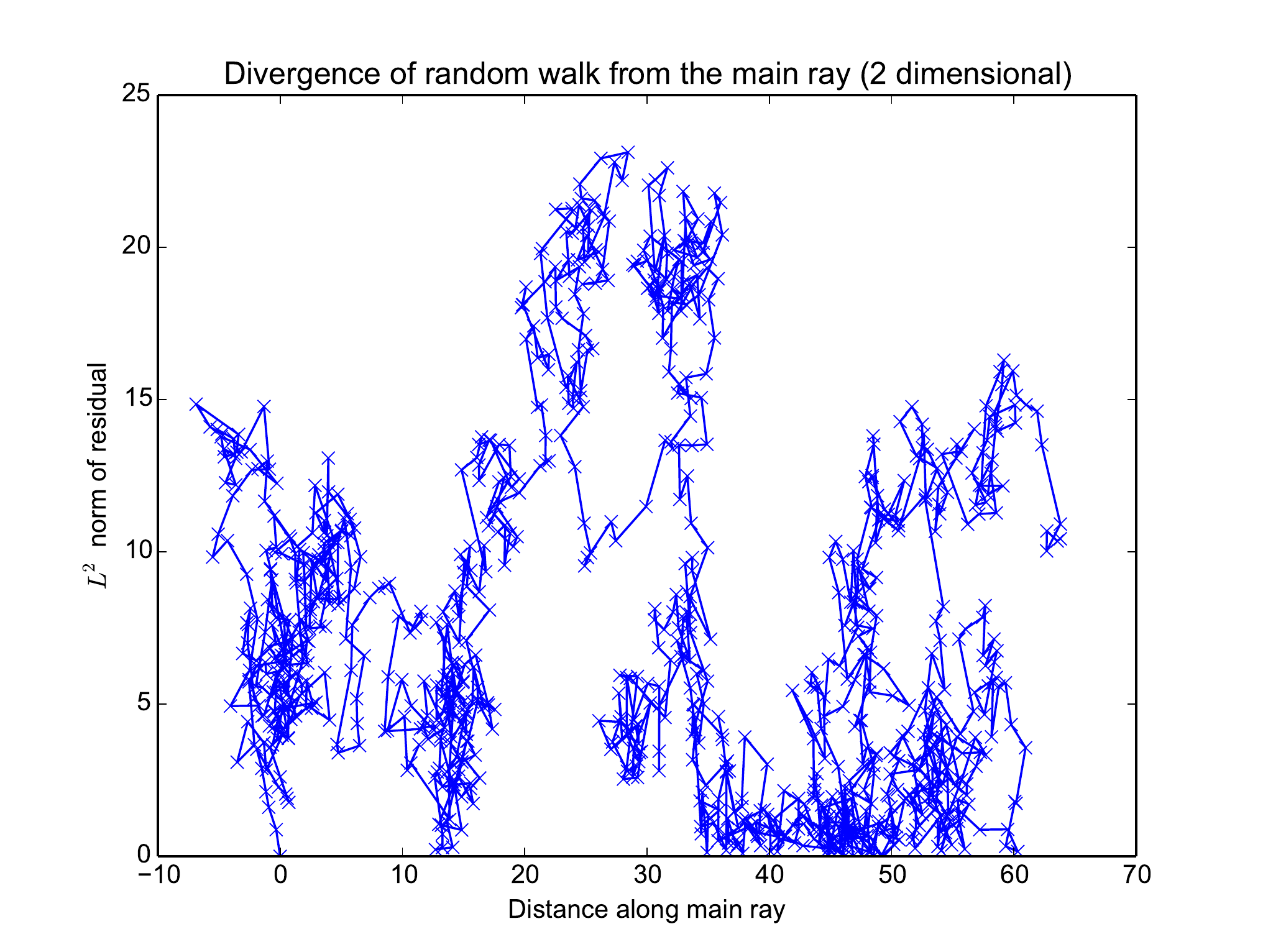} &%
    \includegraphics[width=.4\textwidth]{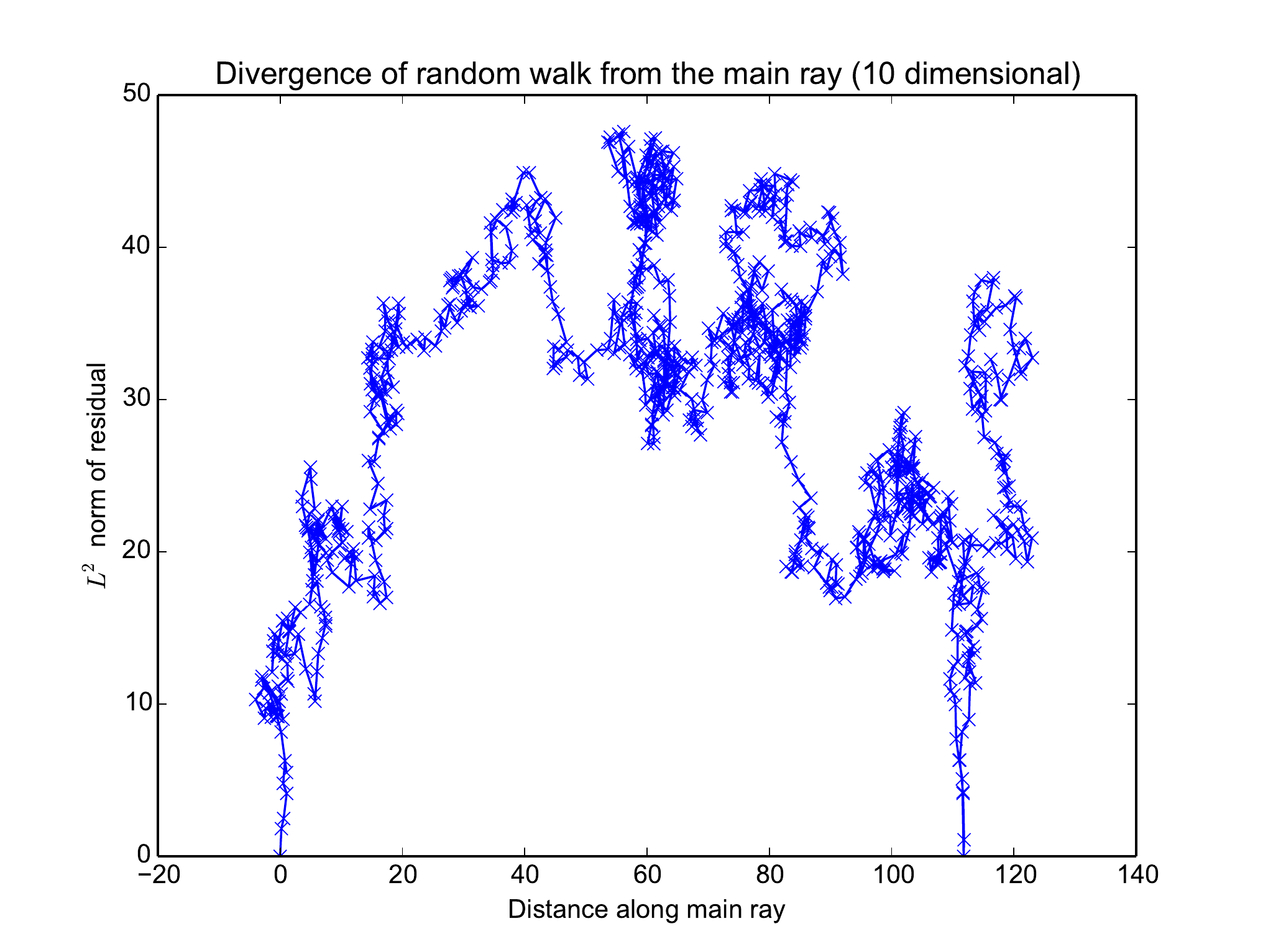} \\
    \includegraphics[width=.4\textwidth]{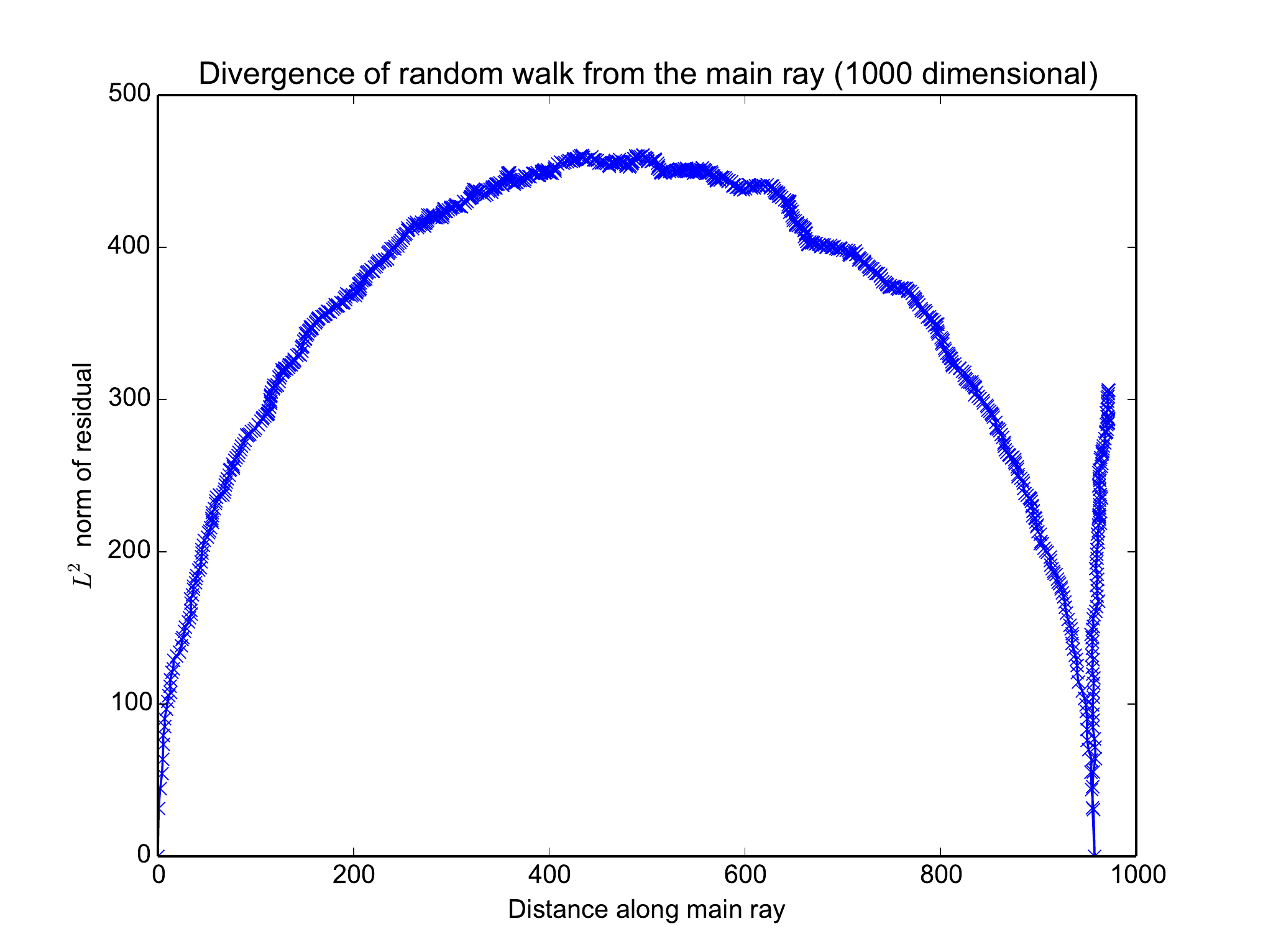} &%
    \includegraphics[width=.4\textwidth]{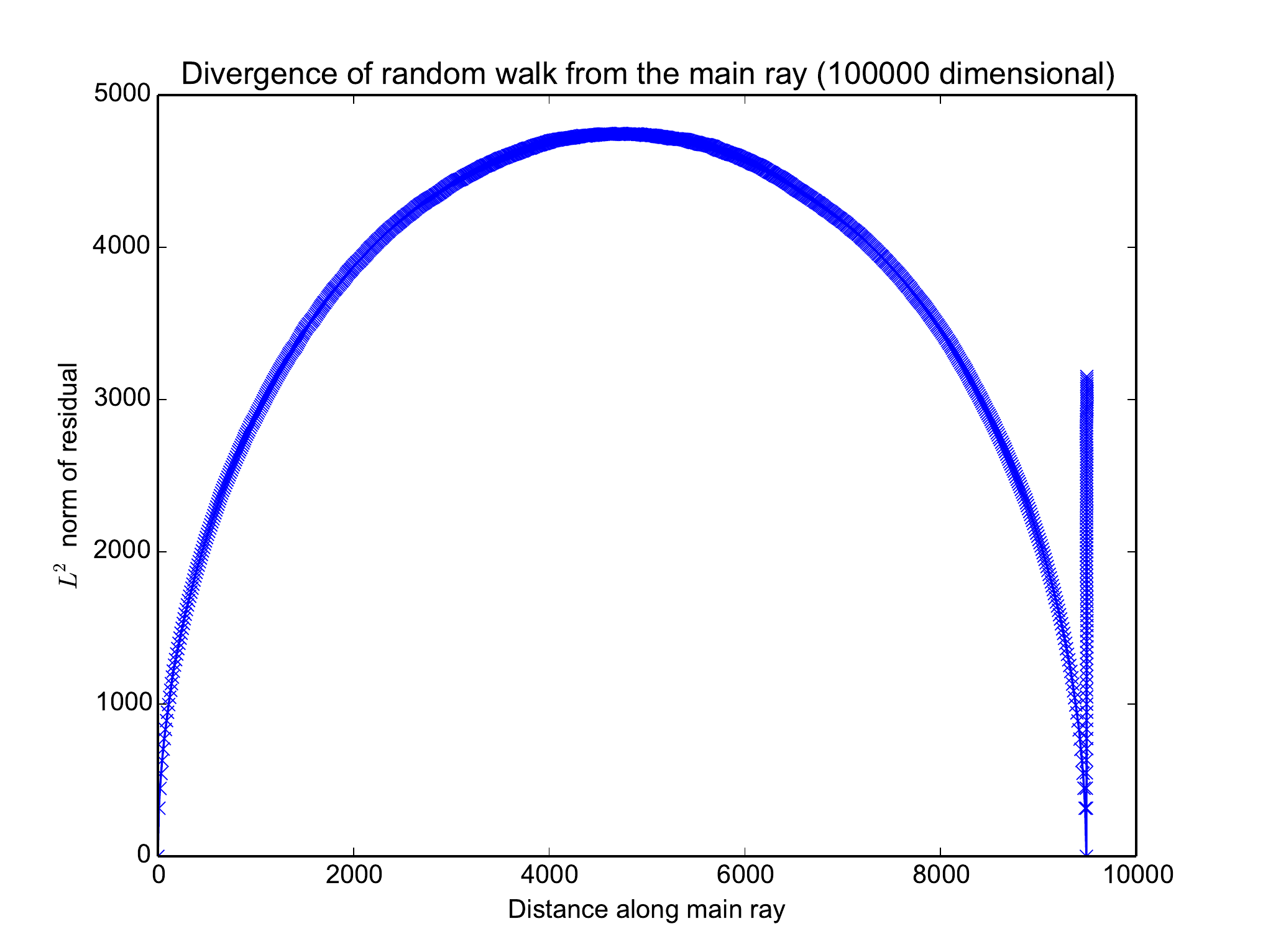}
  \end{tabular}
  \caption{Plots of the projection along the axis from initialization to solution
    versus the norm of the residual of this projection for random walks of varying
    dimension. Each plot is formed by using 1,000 steps. We designate step 900
    as being the ``solution'' and continue to plot 100 more steps, in order to
    simulate the way neural network training trajectories continue past the
    point that early stopping on a validation set criterion chooses as
    being the solution. Each step is made by incrementing the current coordinate
    by a sample from a Gaussian distribution with zero mean and unit covariance.
    Because the dimensionality of the space forces most trajectories to have
    this highly regular shape, this kind of plot is not a meaningful way of
    investigating how SGD behaves as it moves away from the 1-D subspace we
    study in this paper.
  }
  \label{fig:murkwood_gaussian}
\end{figure}

In the next section, we will see that the shape of the objective function in terms of
$\alpha$ and $\beta$ coordinates has interesting features that vary from one problem
to another. However, it is important to understand that this kind of plot does not
tell us much about the shape of the SGD trajectory. All that it tells us is how far
SGD strays from the primary linear subspace.
This is because in high dimensional spaces, the shape of
this curve does not convey very much information. See Fig.~\ref{fig:murkwood_gaussian}
for a demonstration of how this plot converges to a simple geometric shape as the
dimensionality of a random walk increases.

\begin{figure}
  \centering
  \includegraphics[width=.4\textwidth]{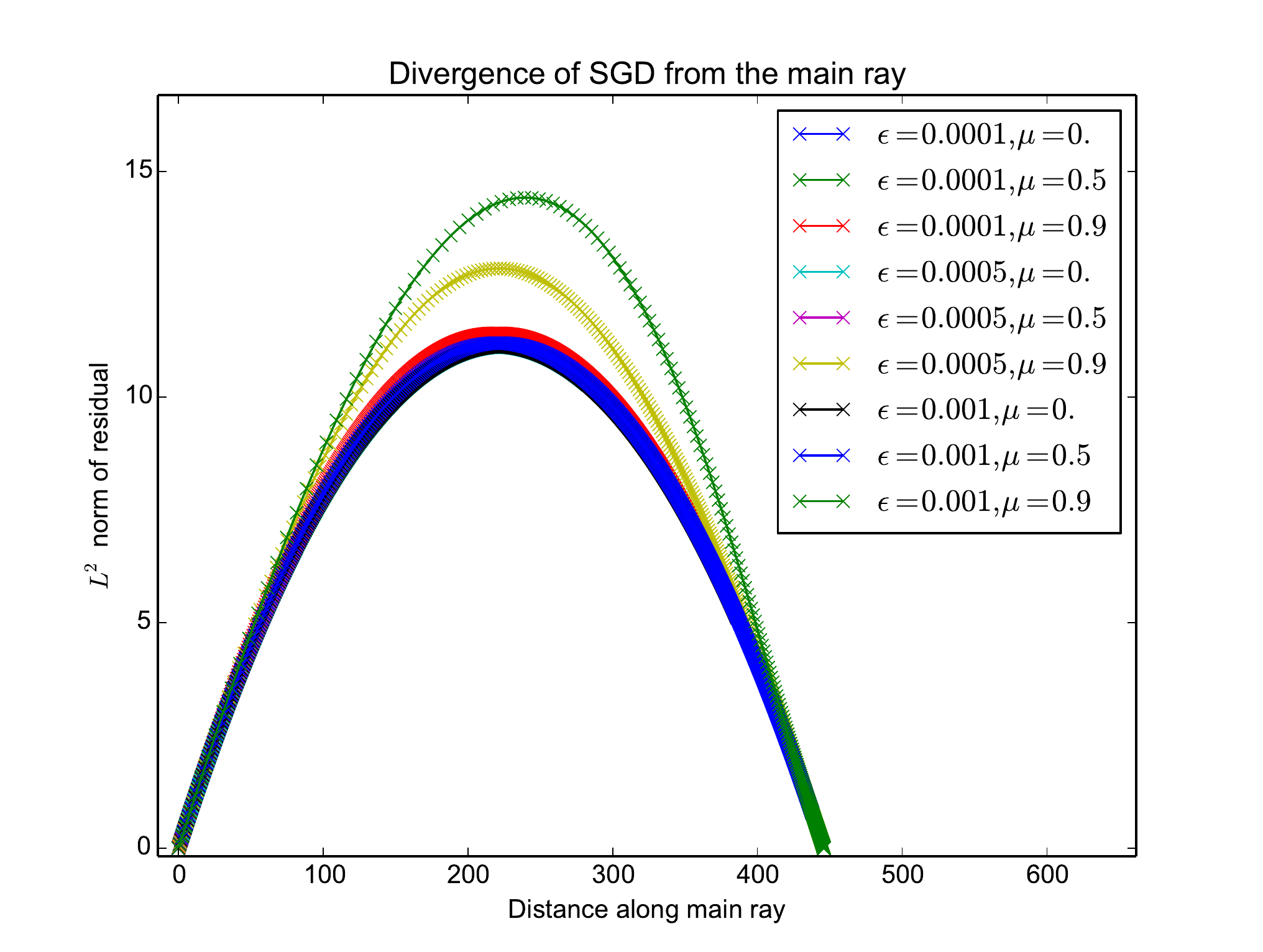}
  \caption{To show the effect of different learning rates $\epsilon$ and
    momentum coefficients $\mu$, we plot the projection and residual norm
    of gradient descent with several different hyperparameters. In this case,
    to make the plots comparable, we use the true solution to a synthetic,
    convex problem as the endpoint for all trajectories. (In the non-convex case,
    different trajectories could go to different solutions, and this would confound
    our interpretation of the differences between them) Because this problem
    is 100,000 dimensional, the curves all have a very simple shape, and the
    primary quantity distinguishing them is the maximum norm of the residual.
   }
   \label{fig:murkwood_quadratic}
 \end{figure}

Plots of the residual norm of the projection for SGD trajectories
converge to a very similar geometric shape in high dimensional spaces.
See Fig.~\ref{fig:murkwood_quadratic} for an example of several different runs of
SGD on the same problem. However, we can still glean some information from this
kind of plot by looking at the maximum norm of the residual and comparing this to
the maximum norm of the parameter vector as a whole.

\begin{figure}
\centering
\includegraphics[width=.4 \textwidth]{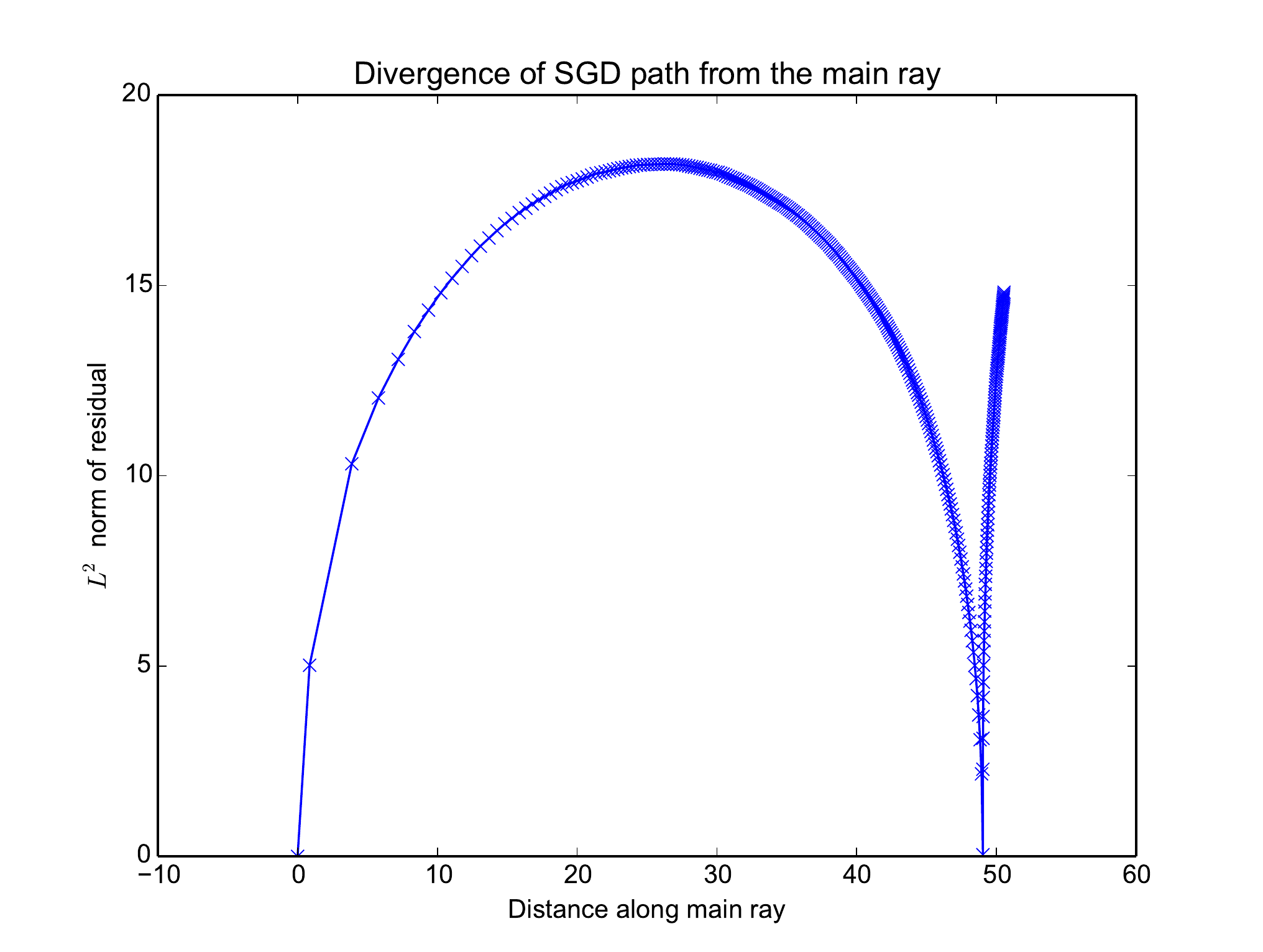}
\caption{This plot examines how far off the linear path SGD strays when training a maxout
  network on MNIST. The x axis is
the projection along the linear path from initialization to solution. The y axis is
the norm of the residual. The plot uses the $L^p$ norm with $p=2$, also known as the Euclidean norm.
It is not the squared norm.
}
\label{fig:murkwood}
\end{figure}

We show this same kind of plot for a maxout network in Fig.~\ref{fig:murkwood}.
Keep in mind that the shape of the trajectory is not interesting, but the ratio
of the norm of the residual to the total norm of the parameter vector at each point
does give us some idea of how much information the 1-D projection discards. We see
from this plot that our linear subspace captures at least 2/3 the norm of the parameter vector
at all points in time.

\section{Three-dimensional visualizations}

A natural question is whether there exist obstacles in between the well-behaved linear subspace
we have described and the path followed by SGD. One way to investigate this is to introduce an
additional direction of exploration. Specifically, we would like to explore the line passing
through each SGD point and its projection on the primary 1-D subspace we have investigated so far.
If this line contains obstacles, it could explain why SGD does not exploit the simple behavior
within this subspace.

\begin{figure}
  \centering
  \includegraphics[width=\textwidth]{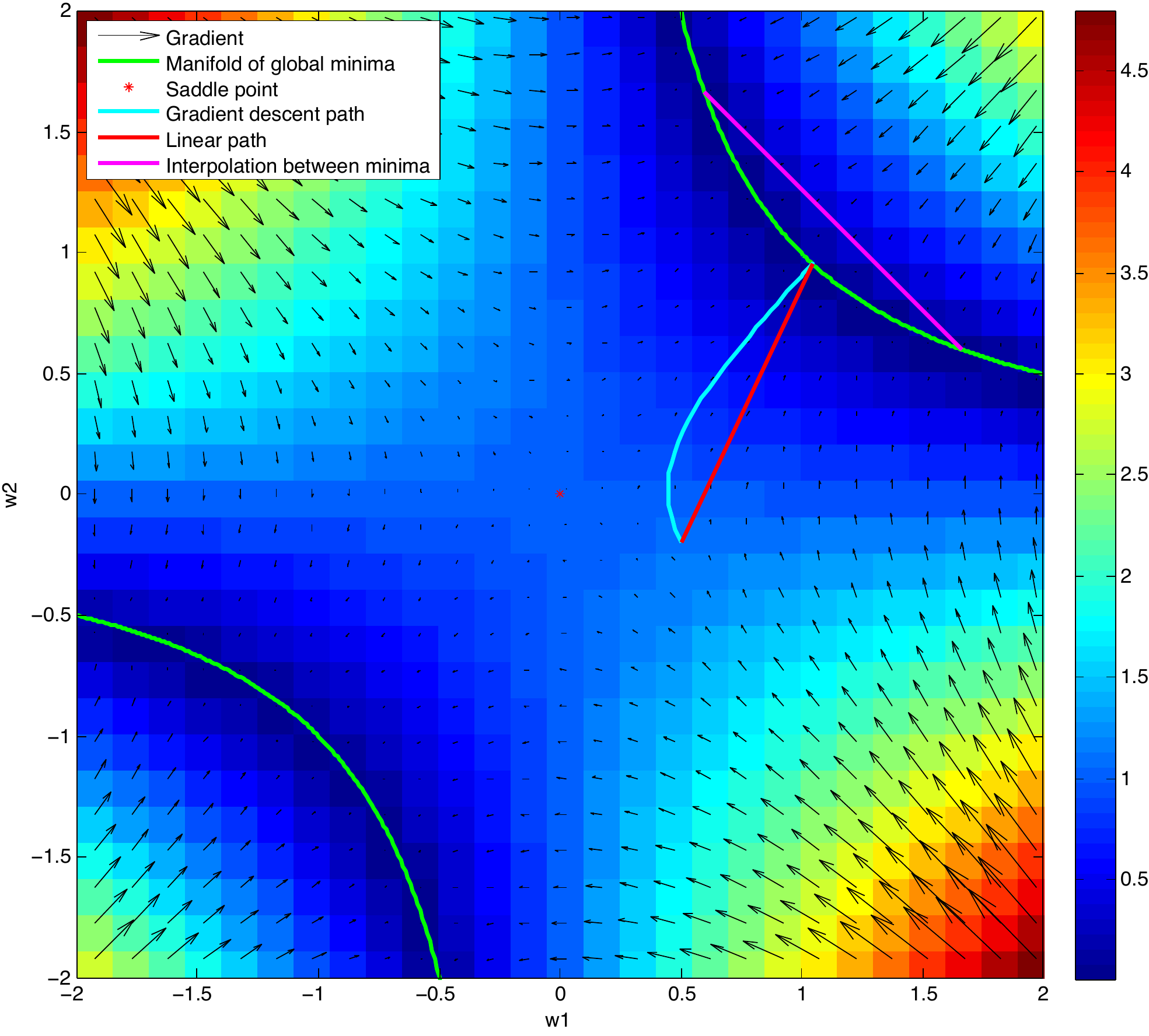}
  \caption{The error surface $(1 - w_1 w_2)^2$ of a factored linear model with two layers and one unit per layer.
    This error surface shows a saddle point at the origin and a non-linear manifold of global solutions.
    The SGD trajectory from initialization to solution encounters negative curvature near its initial point
    and positive curvature near its final point, but does not encounter any exotic obstacles.
  }
  \label{fig:heatmap}
\end{figure}

For our factored linear model, where the training loss is just $(1 - w_1 w_2)^2$, we can accomplish
this by viewing a heatmap of the cost function. See Fig~\ref{fig:heatmap}. This figure predicts
many properties that we expect to see for our more complicated neural network models: negative
curvature at initialization, positive curvature surrounding the solution, a lack of obstacles
separating the SGD trajectory from the well-behaved region of the function, and a connected manifold
of globally optimal points. In this case, the connected manifold is the hyperbola $w_2 = 1 / w_1$.
For neural networks, the pattern of equivalent solutions will be different. For example, we
can take a deep rectified linear network and obtain an equivalent network by rescaling its parameters.
If we modify the parameters of layer $i$ by multiplying bias $j$ and column $j$ of the weight matrix
by $\gamma$, then we can preserve the function of the input that the network respesents by dividing
row $j$ of the weights for layer $i + 1$ by $\gamma$. In the case of the factored linear model, the hyperbolic shape of this manifold means that linearly interpolating between two solution points reveals a region of high cost in the middle, even though the two solutions are
connected by a manifold of other solutions.

This kind of 3-D plot is not directly achievable for problems with more than two parameters. We must
summarize the parameters with a 2-D coordinate somehow. In the remainder of this section, we summarize
the parameters via their projection in to the line spanning initialization and solution (the $\alpha(t)$
coordinate defined in the previous section) and their projection into the line orthogonal to this subspace
that passes through the SGD point $\vtheta(t)$, which we defined as the coordinate $\beta(t)$ in the
previous section.

\begin{figure}
  \includegraphics[width=\textwidth]{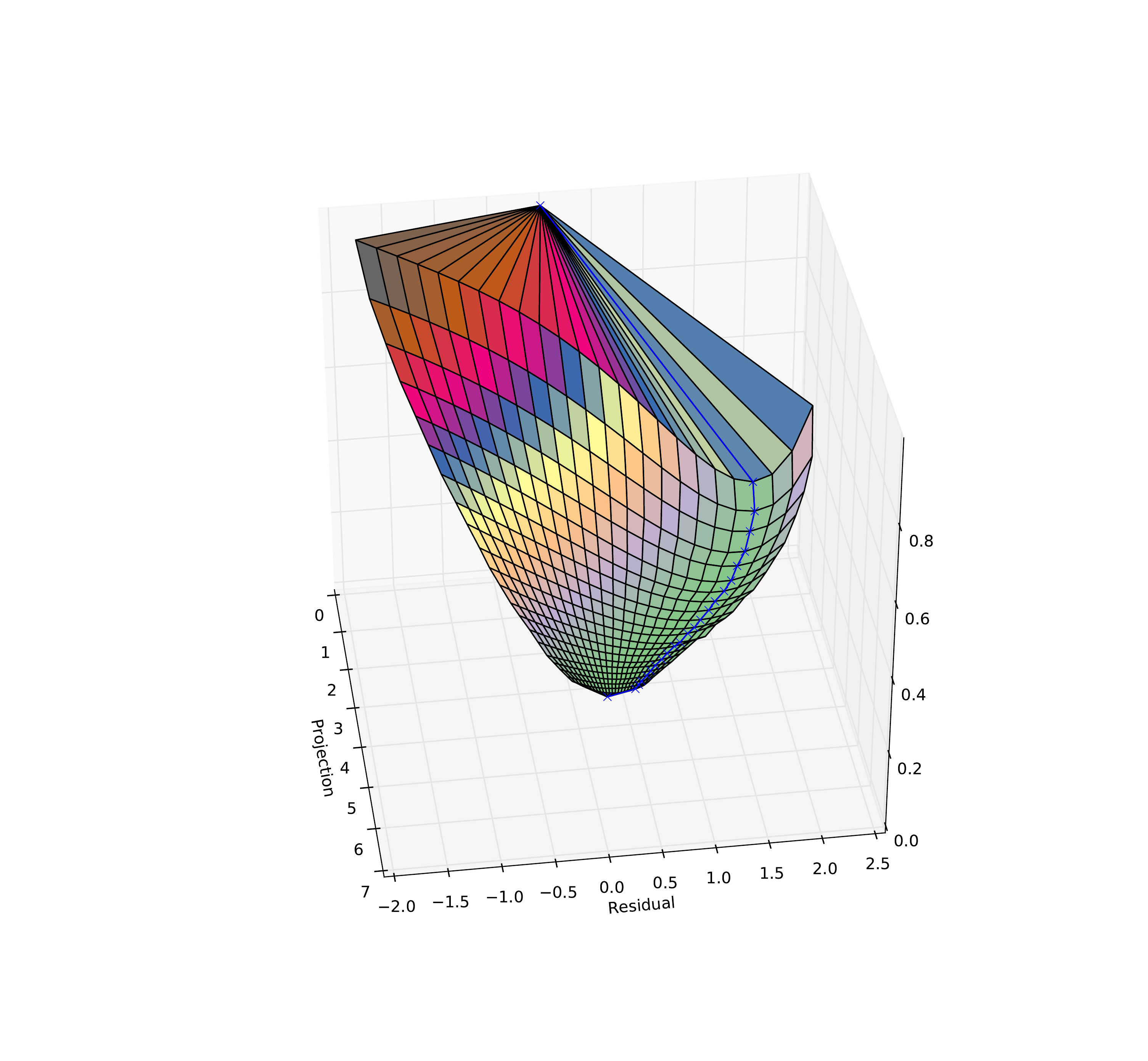}
  \caption{As a canonical example of the deep factored linear model of deep network training, we trained
    a linear model with mean squared error on MNIST. Specifically, we multiply together ten matrices,
    the first nine of which are square and have the same dimension as the MNIST input, and the last of
    which has only ten columns. The output is thus a ten dimensional vector. The mean squared error encourages
    element $i$ of this vector to have value close to 1 and the other elements to have zero when the true
    class is $i$. This 3-D plot shows negative curvature near the initialization point, positive curvature
  near the solution point, and a general lack of obstacles.}
  \label{fig:linear_3d}
\end{figure}

\begin{figure}
  \includegraphics[width=\textwidth]{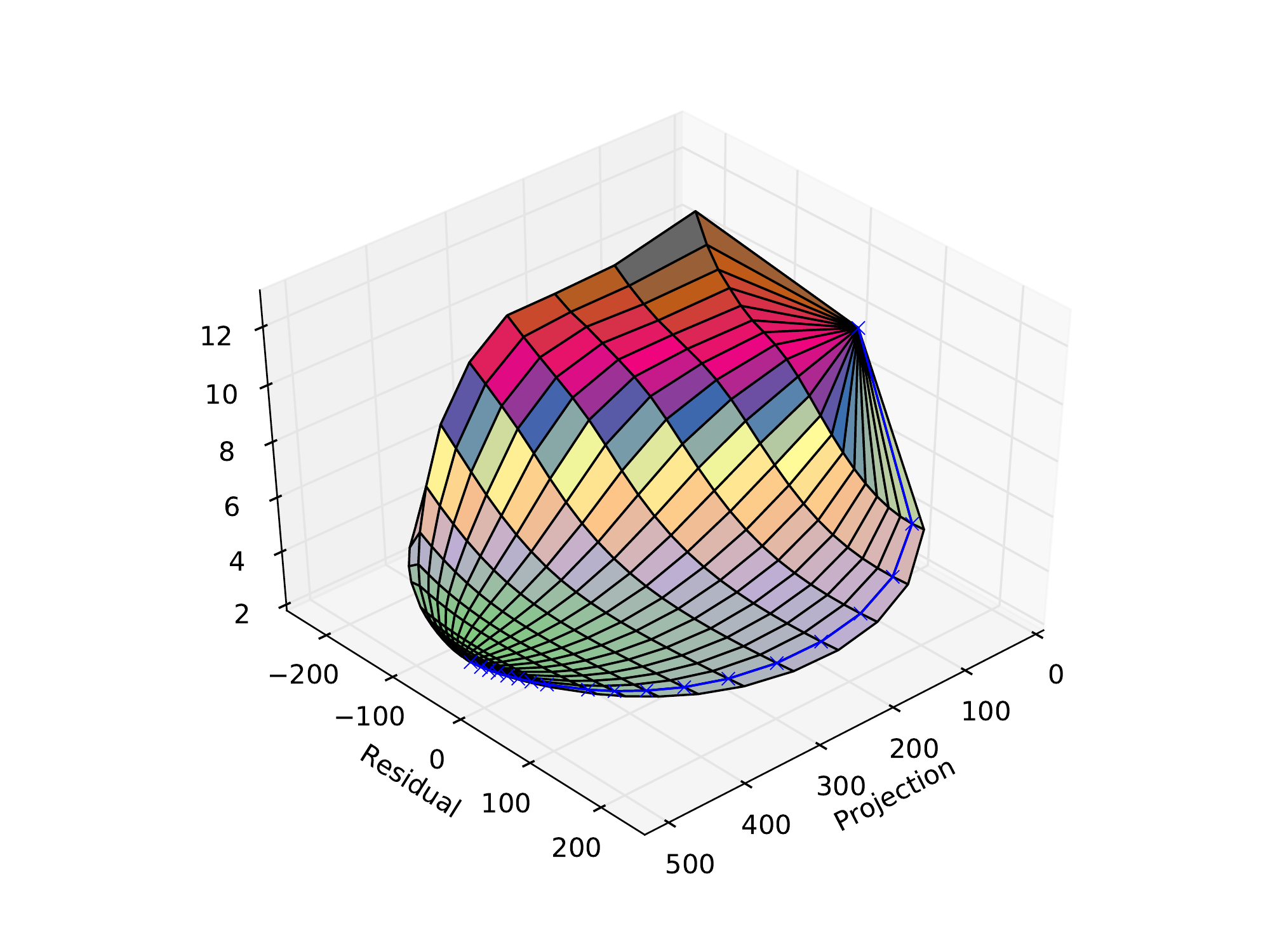}
  \caption{The 3-D visualization of the LSTM cost reveals a simple structure, qualitatively the same
  as that of the deep factored linear model.}
  \label{fig:lstm_3d}
\end{figure}

\begin{figure}
  \includegraphics[width=\textwidth]{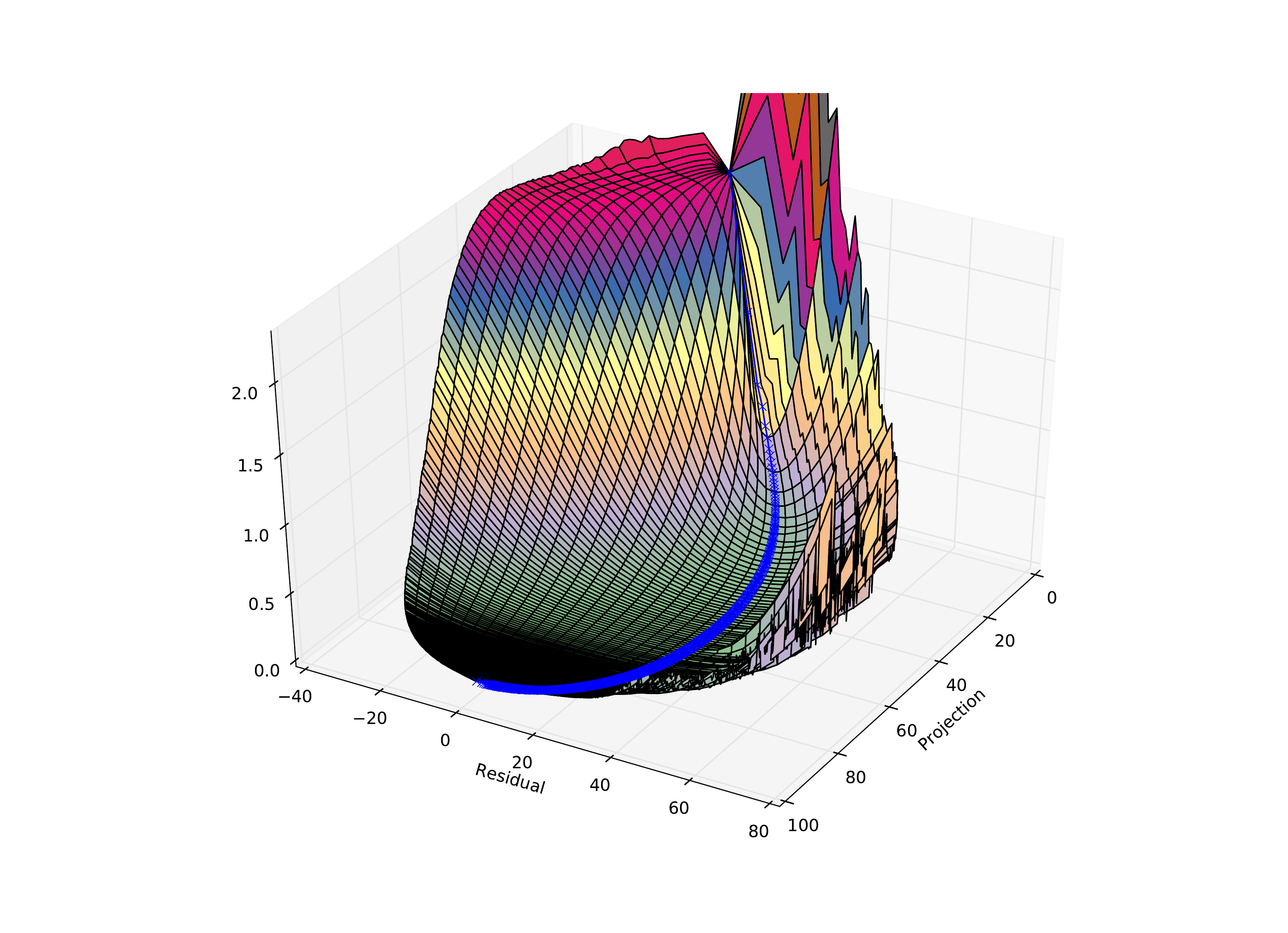}
  \caption{Most of our 3-D visualizations feedforward networks had shapes that were qualitatively the same
    as the factored linear network. Here we show the adversarially trained ReLU network
  as a representative sample.}
  \label{fig:relu_3d}
\end{figure}

\begin{figure}
  \includegraphics[width=\textwidth]{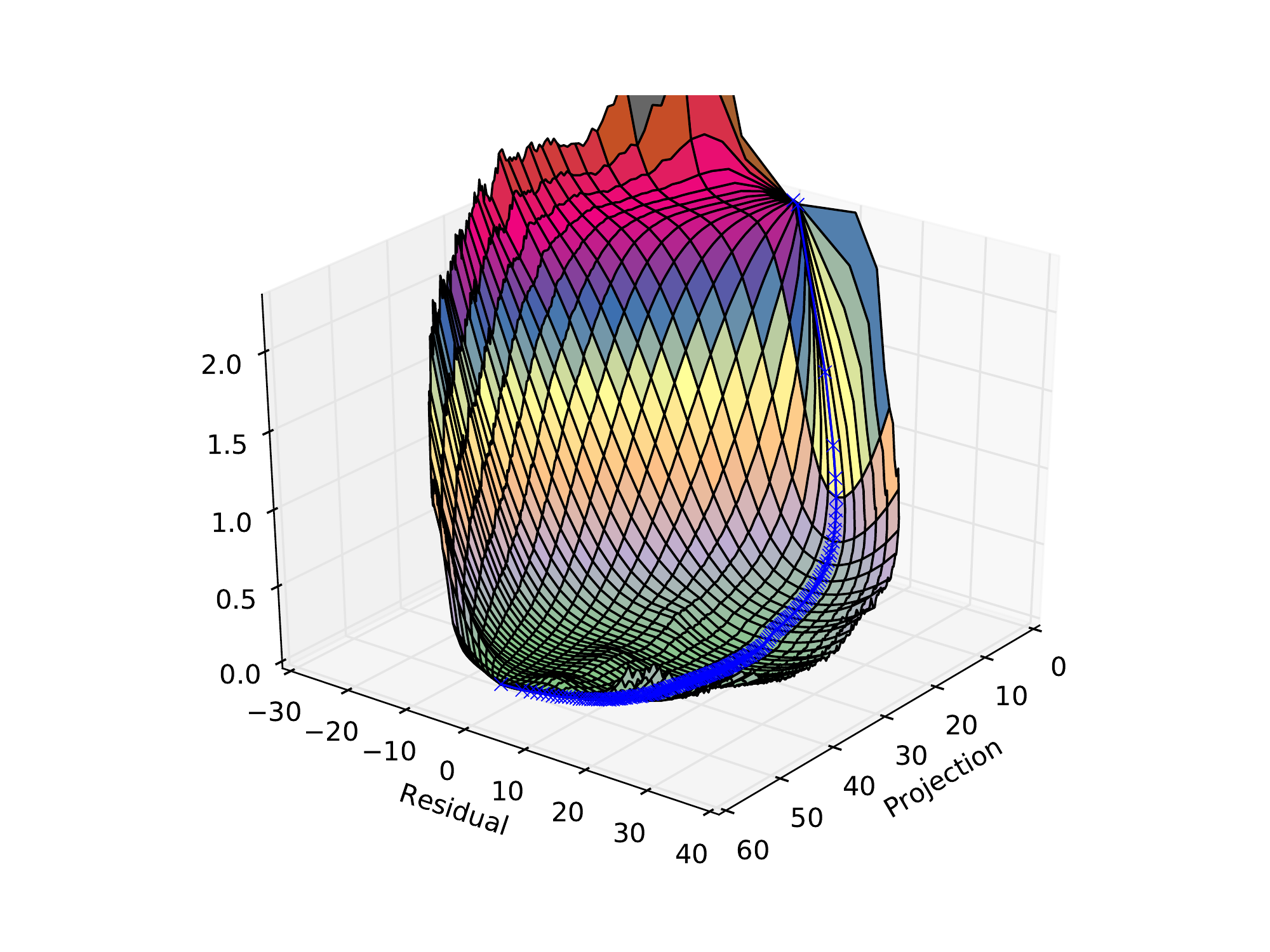}
  \caption{The 3-D visualization technique applied to adversarially trained maxout reveals some obstacles.}
  \label{fig:maxout_3d}
\end{figure}

\begin{figure}
\includegraphics[width=\textwidth]{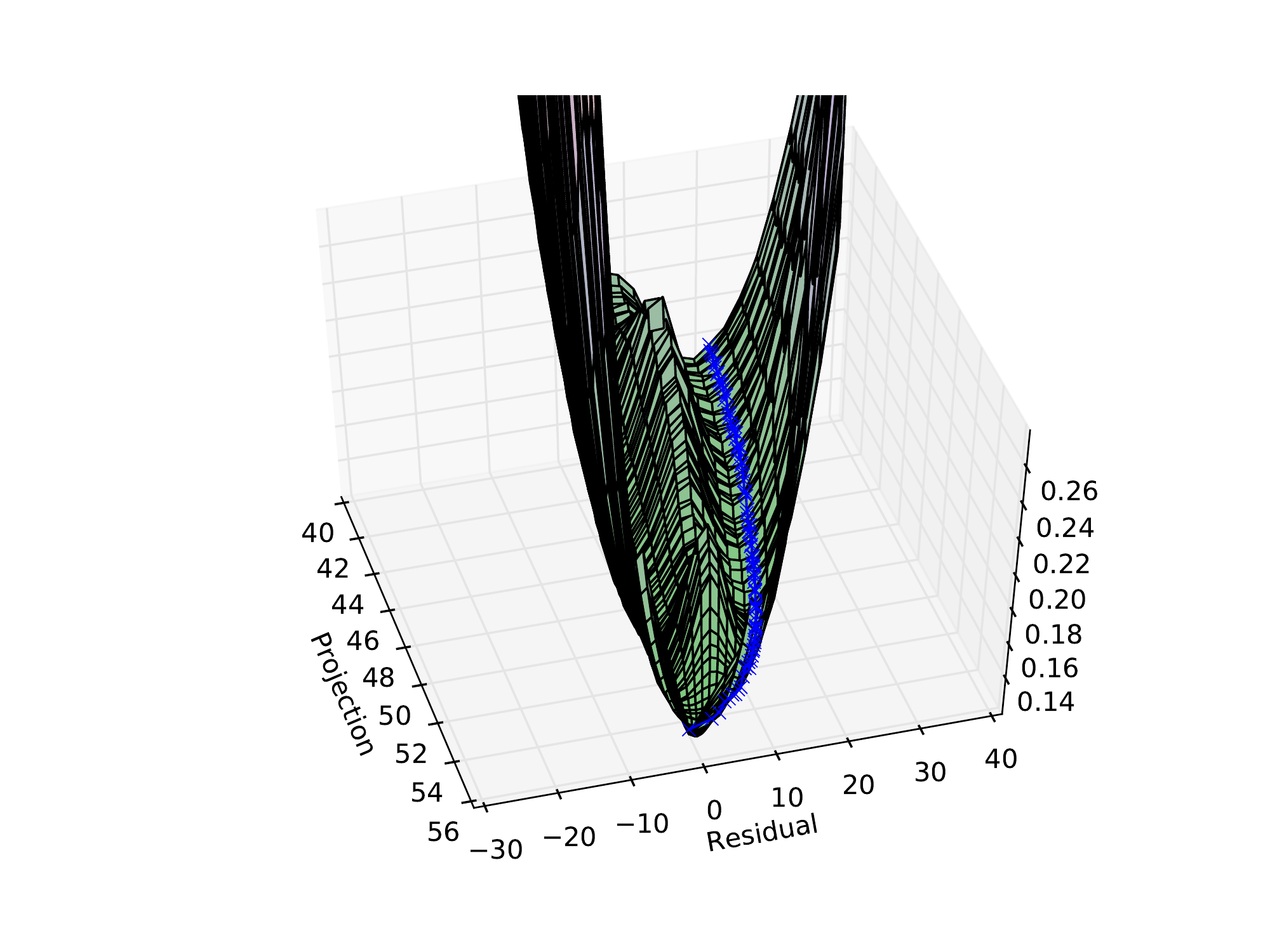}
\caption{The same as Fig.~\ref{fig:maxout_3d}, but zoomed in to show detail near the end of learning.}
\label{fig:maxout_3d_zoom}
\end{figure}

\begin{figure}
  \includegraphics[width=\textwidth]{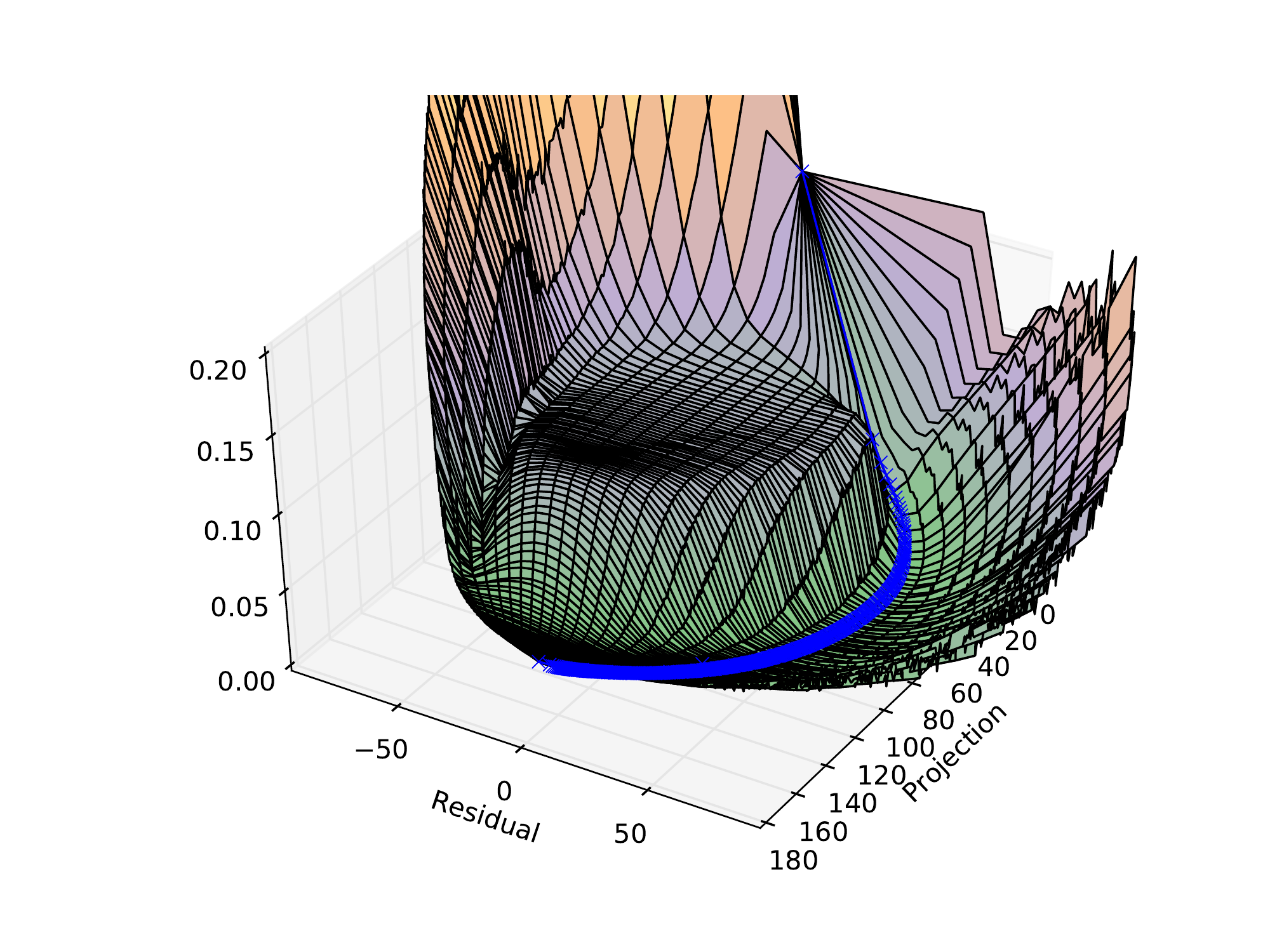}
  \caption{The 3-D visualization of the MP-DBM contains a plateau, but SGD avoided it.}
  \label{fig:mpdbm_3d}
\end{figure}

If we plot the cost as a function of $\alpha$ and $\beta$, we see that the cost function of a deep factored linear model
(Fig.~\ref{fig:linear_3d}) has roughly the same structure as the cost function of our LSTM (Fig.~\ref{fig:lstm_3d})
and as most of our feedforward networks (we show one in Fig.~\ref{fig:relu_3d}). These models show only structures
predicted by the factored linear model of learning dynamics. However, for the adversarially trained maxout network,
we find that an obstacle that is small in height yet very steep constrains SGD into a narrow canyon, preventing it
from accessing the subspace studied in the main text of this paper (Fig.~\ref{fig:maxout_3d} and Fig.~\ref{fig:maxout_3d_zoom}).
Finally, recall that the MP-DBM had a local minimum within the primary 1-D subspace (which could be a local minimum or
a saddle point in high dimensional space). With our 3-D plot (Fig.~\ref{fig:mpdbm_3d}), we see that SGD passed far around the plateau surrounding
this point.

Note that in most of these visualizations we can see signficant negative curvature in the early part
of the SGD trajectory, and that SGD does not seem to have any difficulty escaping the saddle point near
the origin.
One possible explanation for this behavior is that one model of SGD with sufficiently small step size naturally
avoids saddle points. Consider the SGD trajectory as a function of time, $\vtheta(t)$. As an analytical
model of SGD with small step size, we can consider the continuous-time gradient descent process with
$\frac{d}{dt} \vtheta = -\nabla_\vtheta J(\vtheta)$. If we make a second-order Taylor series expansion in time
\[ \vtheta(t) \approx \vtheta(0) - t \frac{d}{dt} \vtheta(t) + \frac{1}{2} t^2 \frac{d^2}{dt^2} \vtheta(t) \]
it simplifies to
\[ \vtheta(t) \approx \vtheta(0) -t \nabla_{\theta(0)} J(\vtheta(0)) + \frac{1}{2} t^2 \mH(0) \nabla_{\vtheta(0)} J(\vtheta(0)) \]
where $\mH$ is the Hessian matrix of $J(\vtheta(0))$ with respect to $\vtheta(0)$.
This view shows that a {\em second-order approximation in time} of continuous-time gradient descent
incorporates {\em second-order information in space} via the Hessian matrix.
Specifically, the second-order term of the Taylor series expansion is equivalent to {\em ascending}
the gradient of $|| \nabla_{\vtheta} J(\vtheta) ||^2$. In other words, the first-order term says to go
downhill, while the second-order term says to make the gradient get bigger. The latter term encourages
SGD to exploit directions of negative curvature.

\section{Control visualizations}

Visualization has not typically been used as a tool for understanding the structure of neural network
objective functions. This is mostly because neural network objective functions are very high-dimensional
and visualizations are by necessity fairly low dimensional.
In this section, we include a few
``control'' visualizations as a reminder of the need to interpret any low-dimensional visualization
carefully.

\begin{figure}
\centering
\includegraphics[width=\textwidth]{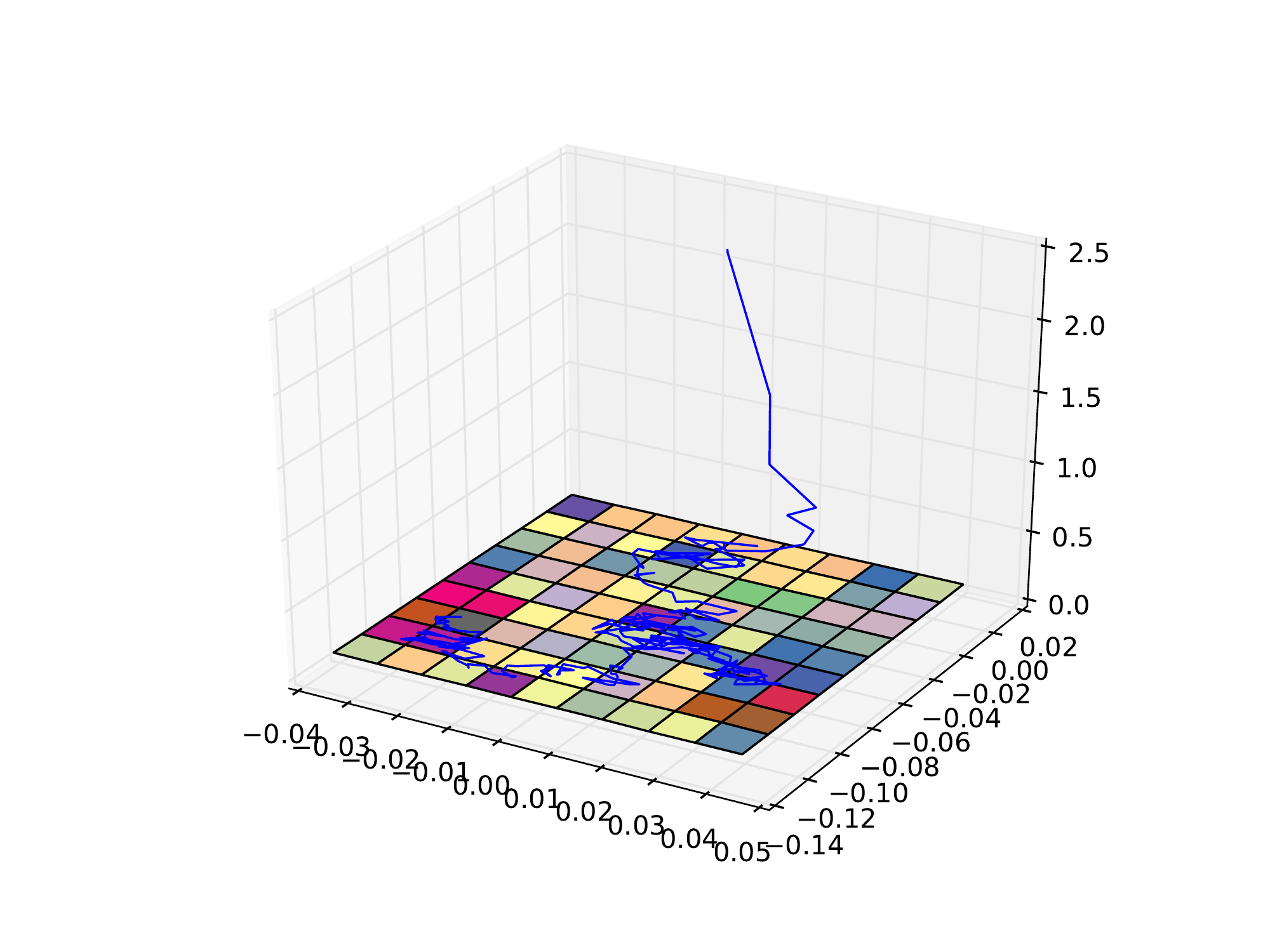}
\caption{As a control experiment, we plot a random 2-D subspace intersecting the solution point.
In this subspace, we see a complicated SGD trajectory and essentially no variation in the cost
function value. This visualization is useful as a reminder that the visualizations presented in
this paper are designed to expose variation in the cost function and discard variation in the
shape of SGD trajectory. Not all directions in the cost function have high variability and
SGD trajectories do vary greatly.
}
\label{fig:random_subspace}
\end{figure}

Most of our visualizations showed rich structure in the cost function and a relatively simple shape
in the SGD trajectory. It's important to remember that our 3-D visualizations are not showing a
2-D linear subspace. Instead, they are showing multiply 1-D subspaces rotated to be parallel to each
other. Our particular choice of subspaces was intended to capture a lot of variation in the cost function,
and as a side effect it discards all variation in a high-dimensional trajectory, reducing most trajectories
to semi-circles. If as a control we instead plot a randomly selected 2-D linear subspace intersecting the
solution point, then we see that there is almost no variation in the cost function within this subspace,
and the SGD trajectory is quite noisy. See Fig.~\ref{fig:random_subspace}.

\begin{figure}
  \centering
  \includegraphics[width=\textwidth]{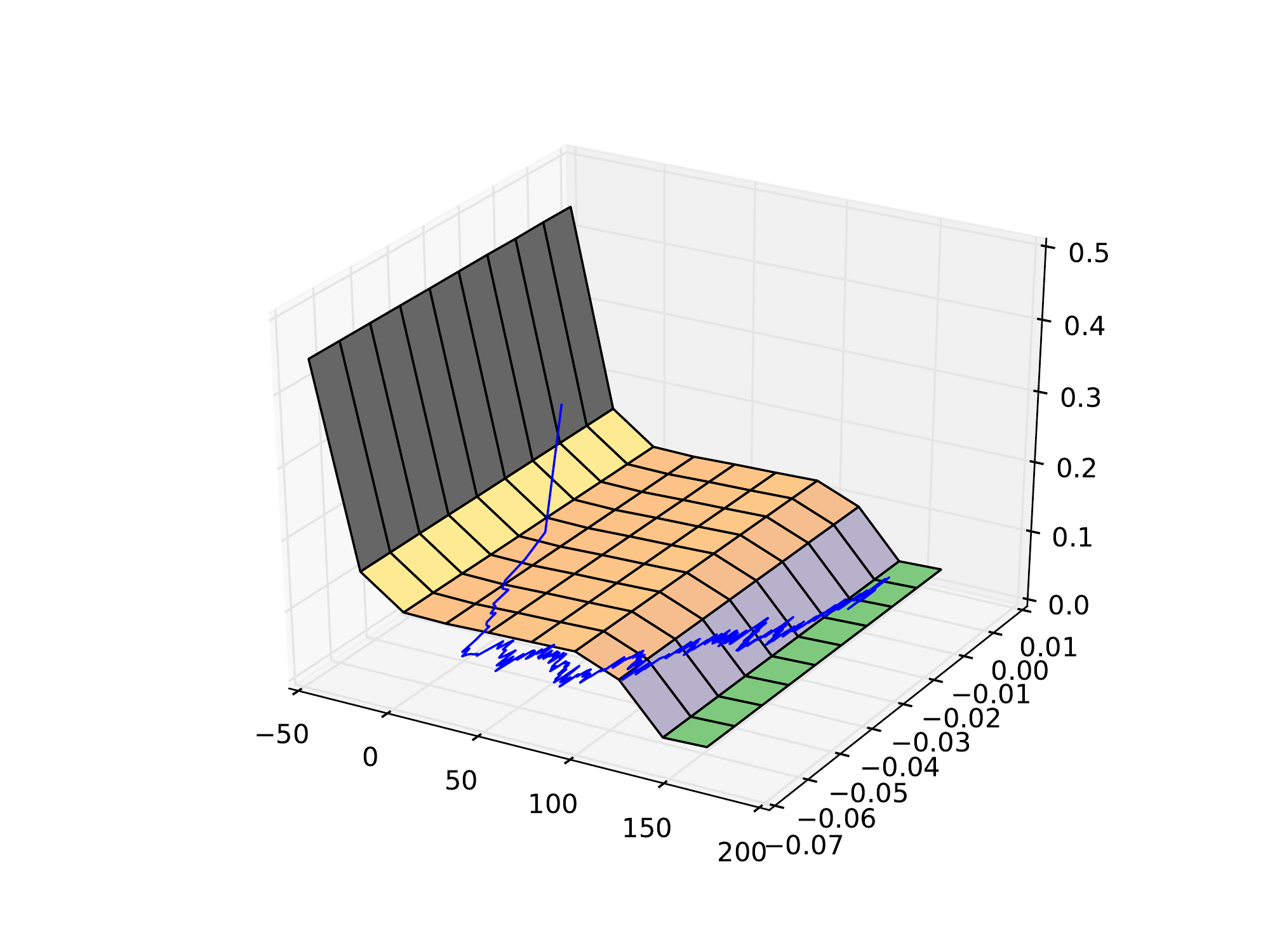}
  \caption{A control experiment with $\alpha(t)$ on one axis and the other axis being a random
  projection.}
  \label{fig:mpdbm_control}
\end{figure}

As an intermediate control, we generated the plot for the MP-DBM, with $\alpha(t)$ on one axis, and the
other axis being a random linear projection. This allows us to see a true 2-D linear subspace that has
significant variation in the cost function due to the choice of the first axis, but also allows us to
see that the SGD trajectory is not a semi-circle. See Fig.~\ref{fig:mpdbm_control}

\end{document}